\documentclass[11pt]{article}
\usepackage{fontspec}
\providecommand{\tightlist}{\setlength{\itemsep}{0pt}\setlength{\parskip}{0pt}}
\usepackage{calc}
\usepackage{fancyvrb}
\usepackage{etoolbox}
\AtBeginEnvironment{verbatim}{\scriptsize}
\usepackage{microtype}
\usepackage[margin=2.5cm]{geometry}
\usepackage{amsmath,amssymb}
\usepackage{booktabs}
\usepackage{longtable}
\usepackage{array}
\usepackage{tikz}
\usepackage{pgfplots}
\pgfplotsset{compat=1.18}
\usepackage{url}
\usepackage[colorlinks=true,linkcolor=black,citecolor=black,urlcolor=blue]{hyperref}
\usetikzlibrary{positioning,arrows.meta,calc,patterns}
\title{KV-LTM: Persistent Per-Session KV Caches as Long-Term Memory for Personal Assistants}
\author{Ziv Kfir\\ \texttt{zivkfir@gmail.com} \and Ido Kfir\\ \texttt{to.ido.kfir@gmail.com}}
\date{}
\begin{document}
\maketitle
\begin{abstract}

\hypertarget{abstract}{%
\label{abstract}}

Personal assistants accumulate months of conversations, but today's
long-term memory stacks put derivation or selection between a question
and its evidence. In a system-level comparison on LongMemEval, two
commercial assistants scored well below the same model reading the
history offline \cite{K1} --- consistent with, but not isolating, content
loss between write and read. We propose KV-LTM: a memory tier storing
each session's raw text plus its KV cache under a fixed reader model,
answering by restoring cached sessions and appending the question ---
complementary to, not a replacement for, retrieval and indexing. Text is
the memory of record; the cache is a recompute-free prefill accelerator,
degrading gracefully to text. We specify the design --- storage record,
budget-derived retention, append-only read path, bounded aggregation. We
size it analytically: across pinned BF16 full-attention (GQA) and
latent-cached MLA configurations, per-token state spans ≈31K--328K bytes
(the pinned KDA+MLA hybrid differs: ≈8.1K bytes/token plus a fixed
per-session recurrent term), giving ≈0.34--3.6 TB per user-year at a
declared 30K-token/day scenario, against ≈49 MB of raw text; per-query
cost is modelled as regime-conditional transfer, attention, decode,
aggregation, and compute. We define an evaluation protocol needing no
persistent-cache implementation for logical QA evaluation: per-session
prefill emulates valid cache reuse under the same model/input-state
contract and ideal arithmetic, with named arms, controlled baselines,
and a preregistered reader-model gate; a separate minimal real-cache
replay tests numerical equivalence and measures the cost axis, outcomes
not assumed. KV-LTM trades storage for avoided re-prefill compute and
write-time fidelity, within a declared retention budget.

\end{abstract}

\hypertarget{chapter-1-introduction-positioning}{%
\section{Introduction \&
Positioning}\label{chapter-1-introduction-positioning}}

\hypertarget{the-personal-assistant-memory-problem}{%
\subsection{The personal-assistant memory
problem}\label{the-personal-assistant-memory-problem}}

A personal assistant (PA) agent accumulates months of interactions:
preferences stated once, decisions made in passing, constraints given
casually. When the user later asks ``what did we decide about X?'', the
answer depends on evidence the system can still reach.

Today's long-term memory (LTM) stacks differ in what they retain,
derive, and select. Extracted-fact memory (e.g., Mem0 \cite{K2}) stores
LLM-distilled facts; temporal graphs (e.g., Zep/Graphiti \cite{K3})
store entities and edges; rolling summaries compress history into
running text --- these transform content at write time. RAG-style
indexing and verbatim archives (e.g., MemPalace \cite{K4}) keep the raw
text but select a subset at read time through a retrieval index;
answering without one means re-reading everything. Both paths cost the
PA, at write time or at read time. KV-LTM targets this trade-off
directly (§1.2).

Public QA measurements show the pattern. On LongMemEval, GPT-4o reading
the history offline scored 0.918 vs 0.577 for ChatGPT(GPT-4o) and 0.330
for Coze(GPT-4o) online with their memory pipelines, on a 97-question
comparison \cite[Fig. 4a]{K1}. On LoCoMo, long-context and RAG systems
remain far behind human accuracy \cite{K5}. On the same benchmark, the
full-context reader scores J≈72.9 vs 66.9 for Mem0 and ≈61 for the best
RAG variant, as reported in Mem0's own (vendor-authored, ECAI 2025)
evaluation under its own protocol \cite{K2}. These are system-level
comparisons --- consistent with, but not isolating, content loss between
write and read (§3 develops the quantitative evidence).

For orientation, we distinguish four memory modes, not an accuracy
ranking or an established taxonomy: \textbf{L1, live context only}, with
no memory retained outside it; \textbf{L2, retrieval or derived memory},
including indexed verbatim archives, extracted facts, graphs and
summaries; \textbf{L3, full-history text reading}, which reads the
entire retained history jointly for each question; and \textbf{L4, the
proposed KV-LTM tier}, which restores per-session native state to avoid
repeated session-text prefill. Systems may combine these modes. Under
the §2.2 contract, a restored session read is equivalent to prefilling
that same session with the same suffix, not to L3's joint attention
across the whole history (§4.7).

In the LongMemEval and Mem0-reported LoCoMo comparisons above, offline
or full-context reading scored higher than the selected memory-system
baselines \cite[Fig. 4a]{K1}\cite{K2}. This is not a universal ordering:
Zep's separate comparison favours its retrieval system (§3.2, §8.1).
KV-LTM targets repeated session-prefill work while retaining original
text; its storage, transfer and read costs are modelled in §5--§7, and
its answer quality remains to be evaluated (§8).

\begin{figure}[!tbp]
\centering
\begin{tikzpicture}
\begin{axis}[
  width=0.48\textwidth, height=5.4cm,
  ybar, bar width=14pt, ymin=0, ymax=1.05,
  ylabel={Accuracy}, ylabel style={font=\small},
  symbolic x coords={Offline read,ChatGPT,Coze},
  xtick=data,
  xticklabels={GPT-4o offline read (L3),ChatGPT (GPT-4o) (L2),Coze (GPT-4o) (L2)},
  x tick label style={font=\scriptsize, rotate=22, anchor=east},
  tick label style={font=\scriptsize},
  nodes near coords, point meta=explicit symbolic,
  nodes near coords style={font=\scriptsize},
  title={(a) LongMemEval --- 97-question subset, 3--6-session histories},
  title style={font=\scriptsize},
]
\addplot coordinates {(Offline read,0.918) [0.918] (ChatGPT,0.577) [0.577] (Coze,0.330) [0.330]};
\end{axis}
\end{tikzpicture}\hfill
\begin{tikzpicture}
\begin{axis}[
  width=0.48\textwidth, height=5.4cm,
  ybar, bar width=14pt, ymin=0, ymax=100,
  ylabel={J score}, ylabel style={font=\small},
  symbolic x coords={Full-context,Mem0,Best RAG},
  xtick=data,
  xticklabels={Full-context (L3),Mem0 (L2),Best RAG (L2)},
  x tick label style={font=\scriptsize, rotate=22, anchor=east},
  tick label style={font=\scriptsize},
  nodes near coords, point meta=explicit symbolic,
  nodes near coords style={font=\scriptsize},
  title={(b) LoCoMo --- vendor-authored Mem0 evaluation},
  title style={font=\scriptsize},
]
\addplot coordinates {(Full-context,72.9) [72.9] (Mem0,66.9) [66.9] (Best RAG,61) [{$\approx$61}]};
\end{axis}
\end{tikzpicture}

\vspace{4pt}
{\scriptsize \tikz[baseline=-0.65ex]{\draw[pattern=north east lines] (0,0) rectangle (0.32,0.2);}~KV-LTM (L4): not yet measured --- §8 (no bar shown; no value implied).\quad L1 (no memory): not plotted --- no published number.\par}
\caption{Published system-level comparisons under two different benchmarks and
source evaluation protocols: (a) LongMemEval's 97-question subset with
3--6-session histories \cite[Fig. 4a]{K1}; (b) LoCoMo J scores from Mem0's
vendor-authored evaluation \cite{K2}. Full-history reading scores highest
among the selected systems shown in each panel, not across all evidence.
Zep's separate vendor-reported LongMemEval-S comparison favours its
retrieval system over a full-context baseline averaging about 115K tokens
and is not plotted here (§3.2, §8.1). These comparisons do not isolate a
memory mechanism, and scores depend on evaluation labels and judges
(§9.1). L1--L4 are our descriptive memory modes; KV-LTM has no measured
accuracy bar (§8). Costs are modelled separately (§7).}
\label{fig:published-memory-comparisons}
\end{figure}

\hypertarget{thesis-a-persistent-per-session-kv-cache-tier}{%
\subsection{Thesis: a persistent per-session KV-cache
tier}\label{thesis-a-persistent-per-session-kv-cache-tier}}

We propose \textbf{KV-LTM}: a per-session long-term memory tier that
stores, for every session, (i) the raw session text and (ii) the
session's KV cache under a fixed reader model. At query time the system
restores cached sessions and appends the question --- reading the
session's native activation state without re-prefilling its text.

Two properties define the design:

\begin{itemize}
\tightlist
\item
  \textbf{Text is the memory of record.} The raw session text is always
  preserved, in every variant of the design. What is preserved at write
  time is the full session content as source text; the native cache
  state differs by architecture class (full-attention/MLA models keep
  per-token state; hybrid models mix a fixed-size recurrent state with
  per-token MLA state).
\item
  \textbf{KV is a recompute-free prefill accelerator.} The cache removes
  repeated prefill of the session text; §7 shows when that saving
  dominates and when byte transfer does instead. The cache is never the
  only copy: if it is evicted, expired, or invalidated by a model
  change, the system degrades gracefully to reading the original text.
\end{itemize}

\hypertarget{complementary-not-a-replacement}{%
\subsection{Complementary, not a
replacement}\label{complementary-not-a-replacement}}

KV-LTM does not replace indexing, retrieval, or compressed memory. It
complements them in two concrete ways, both developed in §4:

\begin{itemize}
\tightlist
\item
  \textbf{(a) Bounded recent evidence-preserving tier.} For the recent
  retention window, KV-LTM answers by reading retained sessions directly
  --- no retrieval index is involved on this reference path.
\item
  \textbf{(b) Verifier/reranker over retrieval candidates.} Behind an
  existing retrieval system, KV-LTM reads the cached KV of the candidate
  sessions the retriever returns --- verifying and reranking them
  against full session content instead of re-prefilling their text.
\end{itemize}

In both roles, the KV tier changes how evidence is accessed, not its
fidelity relative to an identical full prefill. Reading the full session
avoids dependence on selected or derived views (chunks, facts,
summaries); whether that improves answers is measured in §8. The
reference path requires no index; the optional hybrid variant adds a
pre-load prefilter over text/metadata (§4).

The reference tier defers question-directed interpretation of retained
session content to read time rather than replacing it with a write-time
summary or extracted facts. It does not eliminate compute: canonical
prefill occurs at ingestion, and restore, suffix processing, decoding
and aggregation remain query costs (§7). Its fidelity claim is relative
to the same reader prefilling the same session and suffix under §2.2;
retention, reader errors, bounded evidence outputs and aggregation still
limit end-to-end answers (§8).

On this reference path, the reader searches each retained session
through attention over its native state rather than through a retrieval
index. This is a per-session read, not joint attention over the archive.
Selection still occurs through retention (§4.8), bounded evidence
aggregation (§4.6), the joint tier (§4.7), and the optional prefilter
(§4.5).

We propose per-session reading from persistent KV caches as another
search tool for personal assistants, complementary to indexes, RAG, grep
and generated search code; its practical value is to be evaluated under
§8.

\hypertarget{scope}{%
\subsection{Scope}\label{scope}}

\begin{itemize}
\tightlist
\item
  \textbf{In scope:} text available to the reader model, including
  textual documents the assistant processes.
\item
  \textbf{Out of scope:} any claim that visual, table, or audio meaning
  survives preprocessing into text; such transformations are excluded or
  separately evaluated. Session identification --- how raw interaction
  is segmented into sessions --- is a separate problem and is out of
  scope; we acknowledge it as a dependency (§9).
\end{itemize}

\hypertarget{contributions}{%
\subsection{Contributions}\label{contributions}}

\begin{enumerate}
\def\labelenumi{\arabic{enumi}.}
\tightlist
\item
  \textbf{Design} (§4): the KV-LTM storage record, budget-derived
  retention policy, read protocol with append-only query path, and a
  bounded aggregation contract.
\item
  \textbf{Feasibility cost model} (§5--§7): per-token KV size across
  pinned model configurations, per-user annual storage envelopes under
  declared volume assumptions, and a regime-conditional per-query
  compute model.
\item
  \textbf{Evaluation protocol} (§8): a plan to measure the KV-LTM
  logical read path on public LTM benchmarks \textbf{without
  implementing persistent caching}, using per-session prefill as an
  exact logical stand-in under ideal arithmetic, with preregistered
  reader-selection gates and controlled baselines.
\end{enumerate}

The remainder: §2 fixes the technical background and positions prior
work; §3 documents where today's methods lose content; §4 gives the
design; §5--§7 size it; §8 defines the evaluation; §9 states failure
modes, limitations, and next steps.
\hypertarget{chapter-2-background-related-work}{%
\section{Background \& Related
Work}\label{chapter-2-background-related-work}}

\hypertarget{kv-cache-semantics-what-reuse-means}{%
\subsection{KV-cache semantics: what ``reuse''
means}\label{kv-cache-semantics-what-reuse-means}}

In a causal transformer (in standard full-attention designs), each layer
caches key/value tensors per token --- the KV cache. Computing a
session's cache once and reusing it later avoids re-running prefill over
the session's tokens. Two reuse regimes matter:

\begin{itemize}
\tightlist
\item
  \textbf{Prefix-exact reuse.} The cached tokens form an unmodified
  prefix of the new computation. The new tokens are \emph{appended},
  continuing the cached position indices. This is the only regime KV-LTM
  uses on its reference path.
\item
  \textbf{Non-prefix reuse.} Cached segments are fused, reordered, or
  repositioned: CacheBlend fuses independently computed chunk caches and
  reports restoring quality by selectively recomputing a fraction of the
  KV rather than reusing everything, with TTFT gains in its tested setup
  \cite{K6}; EPIC reuses caches position-independently \cite{K7};
  TurboRAG relies on altered masks/positions plus fine-tuning \cite{K8};
  Prompt Cache reuses modular attention states at schema-reserved
  positions \cite{K9}. These methods change masks, positions, or both,
  so their reuse is not guaranteed identical to a fresh prefill under
  the original sequence's positions and masks; they approximate or
  change the computation. They are related work, not our mechanism.
\end{itemize}

\hypertarget{the-exactness-lemma-scoped}{%
\subsection{The exactness lemma
(scoped)}\label{the-exactness-lemma-scoped}}

\textbf{Statement.} Take a session cached by a fixed model, and a
question. Restoring the cached session and appending the question is
\emph{mathematically identical, under ideal arithmetic}, to a full
prefill of (session ‖ question) --- provided: the same complete token
prefix (including fixed instructions), the same model weights, continued
position indices (cache holds 0..N−1, query takes N..N+Q−1), pinned RoPE
base/scaling/max-position, the same attention masks, and the same native
cache state.

\textbf{Boundaries.} (i) Post-hoc quantization or pruning of the cache
is a distinct variant; a matching dtype label is necessary but not
sufficient for equivalent state evolution. (ii) Real engines use
non-deterministic or batch-variant kernels, so practical equality is
verified with tolerances by a small real-cache replay (§8), not assumed.
(iii) The lemma is per-architecture: exactness under valid reuse gives
no inter-architecture quality guarantee (the RoPE condition is itself
per-architecture --- a hybrid model's MLA layers may be position-free).
We derive the lemma from causal computation and the implementation
contract; no external citation is needed.

\hypertarget{compression-three-classes-and-a-hybrid-row}{%
\subsection{Compression: three classes and a hybrid
row}\label{compression-three-classes-and-a-hybrid-row}}

``KV compression'' names unlike mechanisms; we separate them because the
distinction decides what KV-LTM may claim.

\begin{longtable}[]{@{}
  >{\raggedright\arraybackslash}p{(\columnwidth - 6\tabcolsep) * \real{0.2500}}
  >{\raggedright\arraybackslash}p{(\columnwidth - 6\tabcolsep) * \real{0.2500}}
  >{\raggedright\arraybackslash}p{(\columnwidth - 6\tabcolsep) * \real{0.2500}}
  >{\raggedright\arraybackslash}p{(\columnwidth - 6\tabcolsep) * \real{0.2500}}@{}}
\toprule\noalign{}
\begin{minipage}[b]{\linewidth}\raggedright
Class
\end{minipage} & \begin{minipage}[b]{\linewidth}\raggedright
Mechanism
\end{minipage} & \begin{minipage}[b]{\linewidth}\raggedright
Keeps every token?
\end{minipage} & \begin{minipage}[b]{\linewidth}\raggedright
Status for KV-LTM
\end{minipage} \\
\midrule\noalign{}
\endhead
\bottomrule\noalign{}
\endlastfoot
(i) Architectural footprint & GQA, MLA --- smaller state per token by
design \cite{K10}\cite{K11} & Yes (for that model) & On-thesis; exact for
that model \\
(ii) Quantization & FP8/INT4 KV (KIVI, KVQuant) \cite{K12} & Yes &
\textbf{Numerically lossy}; must be validated per config \\
(iii) Token eviction & SnapKV, StreamingLLM --- drops tokens; degrades
in multi-turn reuse \cite{K13}\cite{K14} & No & Off-thesis as a memory
mechanism \\
Hybrid native row & KDA recurrent layers (fixed-size state) + MLA layers
(per-token state), e.g., Kimi Linear (Kimi-Linear-48B-A3B, config pinned
in §5) \cite{K15} & Partially (recurrent state is fixed-size) & Separate
accounting: sessions × S\_recurrent + tokens × b\_MLA (§5) \\
\end{longtable}

Class (iii) is the compression that contradicts evidence preservation;
class (i) lowers bytes per token (§5) --- the retention budget bounds
the total (§4, §6); class (ii) is a validated-or-not engineering choice.
The hybrid row's recurrent component does not keep per-token state; its
quality is evaluated per checkpoint by the §8 gate, and no universal
MLA-versus-GQA quality ordering is claimed.

\hypertarget{memory-systems-for-agents-public-primaries}{%
\subsection{Memory systems for agents (public
primaries)}\label{memory-systems-for-agents-public-primaries}}

A compact inventory of today's agent memory systems --- what exists and
what each optimizes. What each paradigm \emph{loses}, with cited
quantitative evidence, is developed in §3.

\begin{longtable}[]{@{}
  >{\raggedright\arraybackslash}p{(\columnwidth - 6\tabcolsep) * \real{0.2500}}
  >{\raggedright\arraybackslash}p{(\columnwidth - 6\tabcolsep) * \real{0.2500}}
  >{\raggedright\arraybackslash}p{(\columnwidth - 6\tabcolsep) * \real{0.2500}}
  >{\raggedright\arraybackslash}p{(\columnwidth - 6\tabcolsep) * \real{0.2500}}@{}}
\toprule\noalign{}
\begin{minipage}[b]{\linewidth}\raggedright
System
\end{minipage} & \begin{minipage}[b]{\linewidth}\raggedright
Paradigm
\end{minipage} & \begin{minipage}[b]{\linewidth}\raggedright
What it optimizes
\end{minipage} & \begin{minipage}[b]{\linewidth}\raggedright
Source
\end{minipage} \\
\midrule\noalign{}
\endhead
\bottomrule\noalign{}
\endlastfoot
Mem0 & Extracted facts & Compact, fast personalization & \cite{K2}
(vendor-authored paper, ECAI 2025) \\
Zep / Graphiti & Temporal knowledge graph & Changing relationships,
current state & \cite{K3} (vendor preprint) \\
Letta / MemGPT & OS-style tiered memory & Self-managed paging between
context and storage tiers & \cite{K16} \\
MemoryBank & Fact memory + forgetting curve & Human-like retention and
decay & \cite{K17} \\
MemoryOS & Hierarchical OS-style store & Short/mid/long-term integration
& \cite{K18} \\
MemPalace & Verbatim archive + spatial index & Full verbatim retention &
\cite[project + independent analysis]{K4} \\
OpenViking & Hierarchical evidence store & Progressive L0/L1/L2 loading,
traceable retrieval & \cite{K19} (vendor docs) \\
LangMem & Memory SDK & Framework-integrated semantic/episodic/procedural
memory & \cite{K20} (vendor docs) \\
Cognee & Graph + vector hybrid & Entity/relationship traversal &
\cite{K21} (vendor docs) \\
\end{longtable}

\hypertarget{kv-reuse-prior-art}{%
\subsection{KV-reuse prior art}\label{kv-reuse-prior-art}}

Persisting and reusing KV state is established engineering.
Cache-Augmented Generation (CAG) \cite{K22} precomputes knowledge-prefix
KV and appends queries --- the nearest prefix-exact prior art. Serving
systems built the infrastructure: PagedAttention/vLLM \cite{K23},
RadixAttention/SGLang \cite{K24}, CacheBlend \cite{K6}, CacheGen
\cite{K25}, LMCache \cite{K26}, Mooncake \cite{K27}, MemServe \cite{K28},
EPIC \cite{K7}, TurboRAG \cite{K8}, modular Prompt Cache \cite{K9}, and
SSD-backed Tutti \cite{K29}. Memorizing Transformers used a kNN memory
over internal K/V pairs --- a related but distinct mechanism \cite{K30}.
A recent survey maps the field \cite{K31}.

\textbf{Novelty.} How is KV-LTM different from CAG or a persistent
prompt cache? The \emph{mechanism} --- a persistent prefix cache ---
exists. The mechanistic delta: CAG keeps one static corpus prefix that
must fit one context; KV-LTM keeps many growing per-session caches,
reads them independently, aggregates the answers, and manages them with
a retention policy and a text fallback. The contribution is what wraps
the mechanism: a \textbf{per-session retention/eviction policy} derived
from a PA storage and latency budget, a \textbf{read/aggregation
protocol} (append-only per-session reads, bounded source-aware
aggregation, text fallback), a \textbf{PA-scale cost model}
(bytes/token, TB/year, per-query regimes), and an \textbf{evaluation
protocol} that measures the logical read path on LTM benchmarks without
an implementation. KV-LTM is positioned as policy + protocol + cost
model + evaluation --- not a new memory primitive.
\hypertarget{chapter-3-where-todays-ltm-loses-content}{%
\section{Where Today's LTM Loses
Content}\label{chapter-3-where-todays-ltm-loses-content}}

Chapter 2 inventoried today's agent memory systems (§2.4). This chapter
documents what their paradigms give up, in two separate evidence
classes: a paradigm-level landscape (§3.1) and cited quantitative loss
evidence (§3.2). Throughout, we distinguish three places where evidence
can become unavailable to the reader: \textbf{write-time derivation}
(the system stores something other than the original), \textbf{source
retention} (whether originals are kept at all), and \textbf{query-time
selection} (the reader sees only a chosen subset).

\hypertarget{landscape-what-each-paradigm-optimizes-and-what-it-gives-up}{%
\subsection{Landscape: what each paradigm optimizes and what it
gives
up}\label{landscape-what-each-paradigm-optimizes-and-what-it-gives-up}}

\emph{Method (author analysis):} we characterize each paradigm along the
three axes above, from each system's public paper or documentation
\cite{K5}\cite{K2}\cite{K3}\cite{K16}\cite{K17}\cite{K18}\cite{K4}\cite{K19}\cite{K20}\cite{K21}.
A derivation or selection step marks where evidence can become
unavailable to the reader; its presence is \textbf{not by itself a
measured loss} --- measured numbers are §3.2's job.

\begin{longtable}[]{@{}
  >{\raggedright\arraybackslash}p{(\columnwidth - 8\tabcolsep) * \real{0.2000}}
  >{\raggedright\arraybackslash}p{(\columnwidth - 8\tabcolsep) * \real{0.2000}}
  >{\raggedright\arraybackslash}p{(\columnwidth - 8\tabcolsep) * \real{0.2000}}
  >{\raggedright\arraybackslash}p{(\columnwidth - 8\tabcolsep) * \real{0.2000}}
  >{\raggedright\arraybackslash}p{(\columnwidth - 8\tabcolsep) * \real{0.2000}}@{}}
\toprule\noalign{}
\begin{minipage}[b]{\linewidth}\raggedright
Paradigm
\end{minipage} & \begin{minipage}[b]{\linewidth}\raggedright
Exemplars (public primary)
\end{minipage} & \begin{minipage}[b]{\linewidth}\raggedright
What it optimizes
\end{minipage} & \begin{minipage}[b]{\linewidth}\raggedright
Originals retained?
\end{minipage} & \begin{minipage}[b]{\linewidth}\raggedright
Where content can be given up
\end{minipage} \\
\midrule\noalign{}
\endhead
\bottomrule\noalign{}
\endlastfoot
RAG / vector indexing & RAG baselines evaluated in LoCoMo \cite{K5} &
Read-time selectivity & Yes (configurations that keep raw chunks) &
\textbf{Query-time selection} --- evidence outside the retrieved subset
never reaches the reader \\
Extracted facts & Mem0 \cite{K2} (vendor-authored paper, ECAI 2025) &
Compact, fast personalization & No --- facts replace the dialogue &
\textbf{Write-time derivation} --- wording, rationale, and uncaptured
context do not survive into the store \\
Temporal graphs & Zep / Graphiti \cite{K3} (vendor preprint) & Changing
relationships; current state & Yes --- raw messages retained as episodes
& Derivation + selection --- the semantic entity/edge view is derived;
superseded facts are \textbf{marked invalid by timestamp, not deleted};
retrieval over the derived view mediates what reaches the reader \\
Hierarchical evidence stores & OpenViking \cite{K19} (vendor docs) &
Progressive L0/L1/L2 loading; traceable retrieval & Yes (L2 originals) &
\textbf{Derived} L0/L1 abstracts mediate access --- selection among them
decides what gets read in full \\
Verbatim archives & MemPalace \cite[project + independent analysis]{K4}
& Full verbatim retention & Yes & \textbf{Query-time selection} via
vector retrieval --- headline recall claims are independently questioned
in the cited analysis \\
OS-style tiered memory & Letta / MemGPT \cite{K16}; MemoryOS \cite{K18} &
Self-managed paging across memory tiers & May persist in a deep tier
(documented for MemGPT: recall storage reached via agent-issued search)
& \textbf{Write-time derivation} on eviction/consolidation --- MemGPT:
recursive summaries of evicted messages \cite{K16}; MemoryOS:
segmented-page organization with LLM-summarized segments \cite{K18} \\
Fact memory + decay & MemoryBank \cite{K17} & Human-like retention and
forgetting & Yes --- conversation records kept alongside daily/global
summaries & Summaries derived at write time; decay-driven forgetting of
unprompted content \\
Memory SDKs & LangMem \cite{K20} (vendor docs) & Framework-integrated
semantic/episodic/procedural memory & Not stated in the cited docs &
\textbf{Write-time derivation} --- extraction-based writes into the
store \\
Graph + vector hybrids & Cognee \cite{K21} (vendor docs) &
Entity/relationship traversal & Documents/chunks tracked in a relational
store & \textbf{Write-time derivation} --- LLM extraction builds the
graph at ingestion \\
\end{longtable}

The pattern: every paradigm in this table either derives at write time
or selects at read time (or both). KV-LTM's design (§4) targets exactly
these two points: within the retained session set, its reference path
adds no further write-time textual summarization or extraction and
applies no pre-load retrieval filtering --- it keeps the original
session text and reads the session's native cache state (§2.3). Hybrid
variants carry a derived, finite-size recurrent state, and
reader/aggregation quality is evaluated, not guaranteed (§8, §9).

\hypertarget{quantitative-loss-evidence}{%
\subsection{Quantitative loss
evidence}\label{quantitative-loss-evidence}}

Numbers below are cited, not anecdotal. System-level QA comparisons are
labelled as such: they do not isolate a causal mechanism, and QA
accuracy is not a retrieval-recall measurement.

\textbf{Anchor.} LongMemEval's own analysis documents the mechanism
directly: compressing sessions into summaries or per-round extracted
facts degrades overall QA through information loss \cite[§5.2]{K1}. The
effect is conditional, and we report it as such: the same study finds
that fact-level values \emph{help} multi-session reasoning questions
\cite[Fig. 5]{K1} --- the effect is granularity- and
question-type-dependent, which is exactly why §8 fixes value granularity
and reports per-ability results. The paper's own error analysis is
consistent: ChatGPT's memory records evidence early but often modifies
it when compressing the history, and most Coze errors come from failing
to record information from some sessions \cite{K1}. This is the
write-time derivation path that the extracted-fact and rolling-summary
paradigms in §3.1 share.

\textbf{Co-anchor: a controlled within-pipeline comparison.} A
controlled experiment isolates write-time derivation loss \cite[v4
pinned]{K32}: within one fixed retrieve--rerank--reason pipeline, only the
stored representation is swapped --- LLM-extracted artifacts vs verbatim
conversation chunks --- with model, retriever, reranker, and judge held
constant. Verbatim chunks win by 15.9 points on LoCoMo categories 1--3
(43.9\% vs 28.0\%; 699 questions over 10 conversations) and by 22.0
points on LongMemEval-S (67.4\% vs 45.4\%; 500 questions); six confound
controls reproduce the effect, and accuracy tracks how much source text
survives in the store. The paper's own limits are reported: chunks
abstain worse. Chunks also lead the real-benchmark
multi-hop/multi-session slices; artifacts win only the paper's
synthetic, ontology-aligned multi-hop probe --- a different setup from
LongMemEval's own Fig. 5 finding above (fact granularity helping
multi-session reasoning there), so the two experiments do not share a
direction. Label: single-author preprint, under review; self-comparisons
within the paper; a scoped pipeline finding, not a verdict on all
extraction methods.

\textbf{Further mechanism evidence.} Two further results document loss
under their stated setups: (i) the same preprint's constraint probe,
illustrated by its ``use type hints everywhere'' example (the paper's
illustrative example, not an independently traced released item): over
50 conversations × 50 turns with four planted facts per conversation,
rolling compression every 40 turns, and a gpt-4o-mini answerer, a
rolling summary keeps ``use type hints'' but drops the quantifier ---
summarization scores \textbf{14.0\% exact match (19.0\% recall)}, the
neighbouring RAG store \textbf{89.5\% exact}, and the paper's
quote-grounded artifact pipeline \textbf{93.0\% exact (97.5\% recall)}
\cite[v4, Table 4]{K32} --- same-batch comparators, a self-comparison
within a single-author preprint; (ii) the Knowledge Objects audit ---
one 36.7× compaction pass over 2,000 structured facts makes \textbf{60\%
of stored facts irrecoverable} on fact-dense content, and cascading
compaction over \textbf{three rounds} (5-seed mean) loses \textbf{54\%
of project constraints} \cite{K33}. Both are preprints and are labelled
as such.

\textbf{Supporting system-level comparison.} On the 97-question subset
evaluated across commercial systems, LongMemEval reports ChatGPT and
Coze (both GPT-4o) dropping 37\% and 64\% respectively relative to
offline reading of the same history (0.918 → 0.577 / 0.330) \cite[Fig.
4a]{K1} --- a system-level comparison across products, not a causal
compression estimate.

\textbf{T5 --- loss-evidence table.}

\begin{longtable}[]{@{}
  >{\raggedright\arraybackslash}p{(\columnwidth - 8\tabcolsep) * \real{0.2000}}
  >{\raggedright\arraybackslash}p{(\columnwidth - 8\tabcolsep) * \real{0.2000}}
  >{\raggedright\arraybackslash}p{(\columnwidth - 8\tabcolsep) * \real{0.2000}}
  >{\raggedright\arraybackslash}p{(\columnwidth - 8\tabcolsep) * \real{0.2000}}
  >{\raggedright\arraybackslash}p{(\columnwidth - 8\tabcolsep) * \real{0.2000}}@{}}
\toprule\noalign{}
\begin{minipage}[b]{\linewidth}\raggedright
Loss mode
\end{minipage} & \begin{minipage}[b]{\linewidth}\raggedright
Mechanism
\end{minipage} & \begin{minipage}[b]{\linewidth}\raggedright
Example
\end{minipage} & \begin{minipage}[b]{\linewidth}\raggedright
Number
\end{minipage} & \begin{minipage}[b]{\linewidth}\raggedright
Label
\end{minipage} \\
\midrule\noalign{}
\endhead
\bottomrule\noalign{}
\endlastfoot
System-level QA gap (recursive summarization vs full conversation) & not
isolated --- system-level comparison; QA accuracy is not a
retrieval-recall measurement & Zep paper's DMR benchmark & recursive
summarization 35.3\% (MemGPT-reported) vs full conversation 94.4\%
(Zep-run); different GPT-4-Turbo snapshots & vendor preprint
\cite{K3} \\
System-level QA gap (full context vs memory system vs RAG) & not
isolated --- system-level comparison; QA is not a candidate-recall
metric; reported subset not independently verified (§8 pins its own) &
Mem0 paper on LoCoMo & full-context J≈72.9 vs Mem0 66.9 vs best RAG ≈61
& vendor-authored paper (ECAI 2025) \cite{K2} \\
Write-time derivation (extraction) & controlled ablation --- only the
stored representation varies (model, retriever, reranker, judge fixed) &
Fidelity-Before-Structure ablation & verbatim chunks vs extracted
artifacts: LoCoMo cat 1--3, 43.9\% vs 28.0\% (699 Qs / 10 convs);
LongMemEval-S 67.4\% vs 45.4\% (500 Qs) & preprint (single author, under
review); self-comparison \cite[v4]{K32} \\
Summarization & constraint quantifier dropped (directly supported by the
source) & \cite{K32} constraint probe (illustrative example: ``use type hints
everywhere'') & summarization 14.0\% exact (19.0\% recall); RAG 89.5\%
exact; quote-grounded artifacts 93.0\% exact (97.5\% recall) --- 50
convs × 50 turns, 4 planted facts/conv, compression every 40 turns &
preprint (single author, under review); self-comparison \cite[v4]{K32} \\
Compaction & fact loss per pass; constraint loss on cascade (directly
supported by the source) & Knowledge Objects audit & 36.7× pass over
2,000 structured facts → 60\% of stored facts irrecoverable; three
cascading rounds → 54\% of project constraints & preprint \cite{K33} \\
System-level QA gap (full context vs memory system, small scale) & not
isolated --- system-level comparison; setup-specific & ConvoMem & full
context 70--82\% vs Mem0 30--45\% below 150 conversations &
vendor-authored preprint (Salesforce) \cite{K34} \\
\end{longtable}

\textbf{How to read T5.} The mechanism-level rows (the \cite{K32} ablation and
probe, Knowledge Objects) isolate a mechanism; the three system-level
rows do not. Together they support the article's narrower claim ---
today's pipelines put derivation or selection between question and
evidence, and the gap is measurable --- without claiming that any single
mechanism explains any single system's score.

The ordering is not universal. In Zep's vendor-reported LongMemEval-S
comparison, its system scored 71.2\% versus 60.2\% for full-context
GPT-4o \cite[Table 2]{K3}. The full-context baseline used histories
averaging about 115K tokens; these are whole-history inputs, not the
per-session reads proposed here. This result neither establishes nor
rules out a benefit from per-session cache reuse; it remains a framing
caution in §8.1 and is not plotted in
Figure~\ref{fig:published-memory-comparisons}.
\hypertarget{chapter-4-kv-ltm-design}{%
\section{KV-LTM Design}\label{chapter-4-kv-ltm-design}}

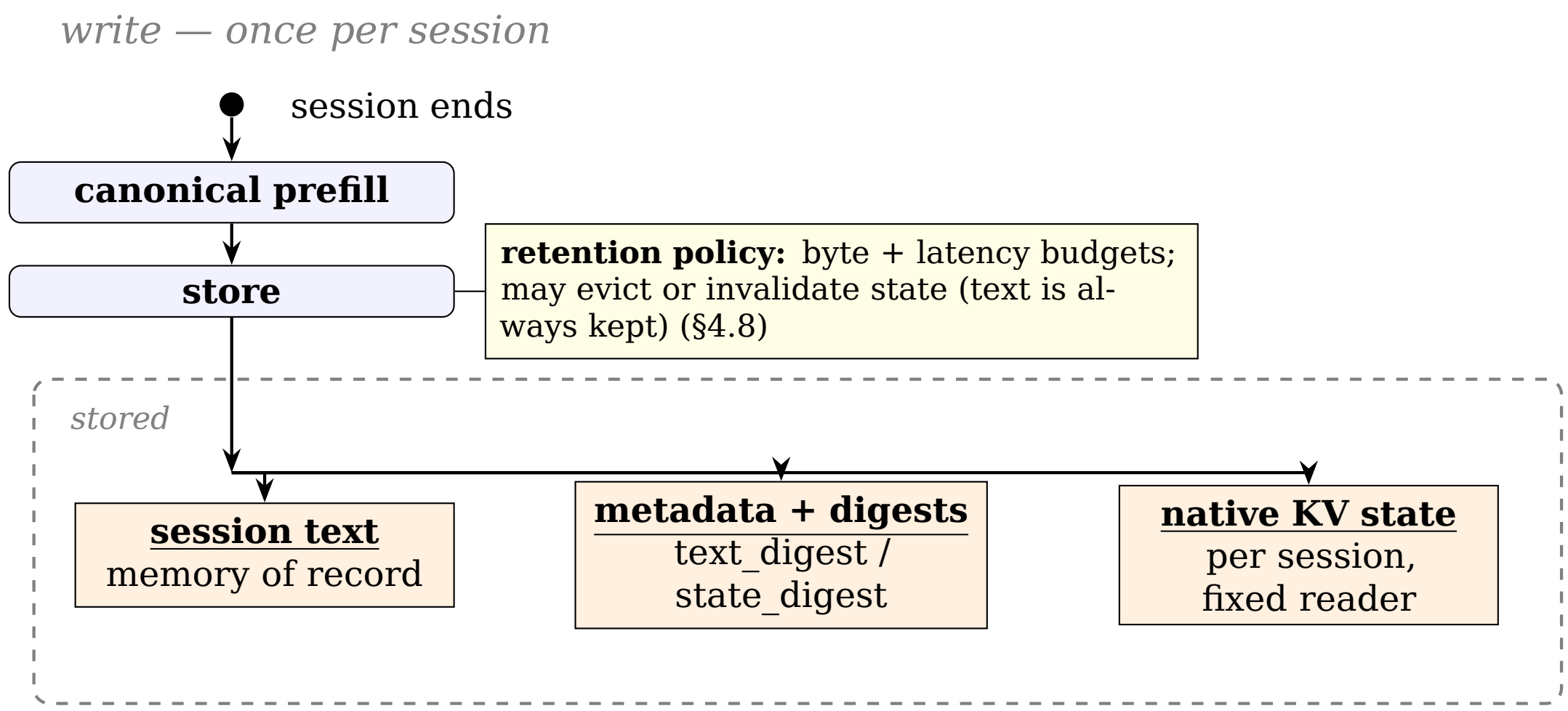
\begin{figure}[t]
\centering
\begin{tikzpicture}[
  font=\footnotesize,
  startnode/.style={circle, fill=black, minimum size=2.2mm, inner sep=0pt},
  action/.style={draw, rounded corners=3pt, align=center, inner sep=3.5pt, fill=blue!5},
  object/.style={draw, sharp corners, align=center, inner sep=3.5pt, fill=orange!12},
  note/.style={draw, sharp corners, align=left, inner sep=4pt, fill=yellow!12, font=\scriptsize},
  flow/.style={-{Stealth[length=2.2mm]}, thick},
  attach/.style={thin},
  bandlab/.style={font=\small\itshape, text=gray},
  region/.style={draw=gray, dashed, rounded corners=4pt, thick},
  rlab/.style={font=\scriptsize\itshape, text=gray}
]
\node[bandlab, anchor=west] at (0.6,0.3) {write --- once per session};
\node[startnode] (start) at (2.3,-0.35) {};
\node[font=\footnotesize, anchor=west] at (2.7,-0.35) {session ends};
\node[action, text width=3.8cm] (prefill) at (2.3,-1.15) {\textbf{canonical prefill}};
\node[action, text width=3.8cm] (store) at (2.3,-2.05) {\textbf{store}};
\draw[flow] (start) -- (prefill);
\draw[flow] (prefill) -- (store);

\node[note, text width=6.2cm] (polnote) at (7.85,-2.05)
  {\textbf{retention policy:} byte + latency budgets;\\ may evict or invalidate state (text is always kept) (§4.8)};
\draw[attach] (store.east) -- (polnote.west);

\draw[region] (0.5,-2.85) rectangle (14.4,-5.8);
\node[rlab, anchor=north west] at (0.7,-2.95) {stored};
\node[object, text width=3.2cm] (otext) at (2.6,-4.45)
  {\underline{\textbf{session text}}\\ memory of record};
\node[object, text width=3.5cm] (ometa) at (7.3,-4.45)
  {\underline{\textbf{metadata + digests}}\\ text\_digest / state\_digest};
\node[object, text width=3.2cm] (ostate) at (12.1,-4.45)
  {\underline{\textbf{native KV state}}\\ per session, fixed reader};
\draw[flow] (store.south) -- (2.3,-3.7);
\draw[flow] (2.3,-3.7) -- (2.6,-3.7) -- (otext.north);
\draw[flow] (2.3,-3.7) -- (7.3,-3.7) -- (ometa.north);
\draw[flow] (2.3,-3.7) -- (12.1,-3.7) -- (ostate.north);

\end{tikzpicture}
\caption{KV-LTM write flow (once per session): canonical per-session prefill creates the stored state at session end; session text, metadata with integrity digests, and native reader state are retained under the §4.8 storage and latency policy, which may evict or invalidate state --- the text is always kept. Boxes denote logical steps or stored objects, not software components.}
\label{fig:write-flow}
\end{figure}

\begin{figure}[!tbp]
\centering
\begin{tikzpicture}[
  font=\footnotesize,
  startnode/.style={circle, fill=black, minimum size=2.2mm, inner sep=0pt},
  endnode/.style={circle, draw=black, thick, double, double distance=1.1pt, fill=black, minimum size=2.0mm, inner sep=0pt},
  action/.style={draw, rounded corners=3pt, align=center, inner sep=3.5pt, fill=blue!5},
  object/.style={draw, sharp corners, align=center, inner sep=3.5pt, fill=orange!12},
  flow/.style={-{Stealth[length=2.2mm]}, thick},
  glab/.style={font=\scriptsize, midway, align=center},
  bandlab/.style={font=\small\itshape, text=gray}
]
\node[bandlab, anchor=west] at (0.6,0.2) {read --- per question (drawing on the stored objects)};

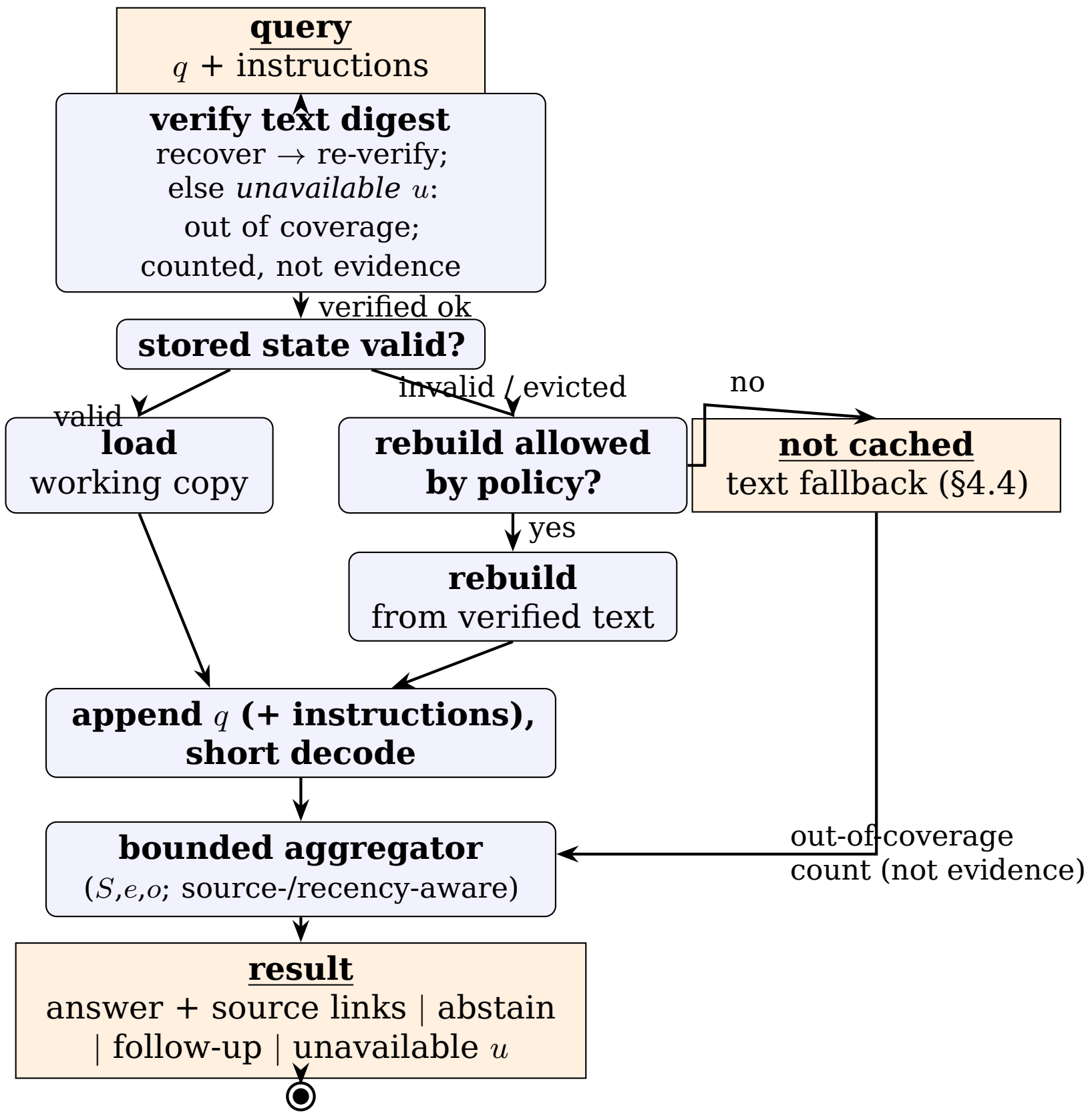
\node[object, text width=3.4cm] (q) at (3.2,-0.85) {\underline{\textbf{query}}\\ $q$ + instructions};
\node[action, text width=4.6cm] (verify) at (3.2,-2.25)
  {\textbf{verify text digest}\\ \scriptsize recover $\rightarrow$ re-verify; else \emph{unavailable $u$}:\\ \scriptsize out of coverage; counted, not evidence};
\node[action, text width=3.4cm] (branch) at (3.2,-3.75) {\textbf{stored state valid?}};
\node[action, text width=2.4cm] (load) at (1.6,-4.95) {\textbf{load}\\ working copy};
\node[action, text width=3.2cm] (gate) at (5.3,-4.95) {\textbf{rebuild allowed}\\ \textbf{by policy?}};
\node[object, text width=3.4cm] (nocache) at (8.9,-4.95)
  {\underline{\textbf{not cached}}\\ text fallback (§4.4)};
\node[action, text width=3.0cm] (rebuild) at (5.3,-6.25) {\textbf{rebuild}\\ from verified text};
\node[action, text width=4.8cm] (decode) at (3.2,-7.6)
  {\textbf{append $q$ (+ instructions),}\\ \textbf{short decode}};
\node[action, text width=4.8cm] (agg) at (3.2,-8.95)
  {\textbf{bounded aggregator}\\ \scriptsize ($S$,$e$,$o$; source-/recency-aware)};
\node[object, text width=5.4cm] (result) at (3.2,-10.35)
  {\underline{\textbf{result}}\\ answer + source links $\vert$ abstain\\ $\vert$ follow-up $\vert$ unavailable $u$};
\node[endnode] (stop) at (3.2,-11.2) {};

\draw[flow] (q) -- (verify);
\draw[flow] (verify) -- node[glab, right=1pt, align=left]{verified ok} (branch);
\draw[flow] (branch.south) ++(-0.7,0) -- (1.6,-4.45) -- (load.north)
  node[glab, pos=0.3, left=0.5pt, align=right]{valid};
\draw[flow] (branch.south) ++(0.7,0) -- (5.3,-4.45) -- (gate.north)
  node[glab, pos=0.22, above=0.5pt]{invalid / evicted};
\draw[flow] (gate.south) -- node[glab, right=0.5pt, align=left]{yes} (rebuild.north);
\draw[flow] (gate.east) -- (7.2,-4.95) -- (7.2,-4.35) -- (nocache.north)
  node[glab, pos=0.25, above=0.5pt]{no};
\draw[flow] (load.south) -- ([xshift=-9mm]decode.north);
\draw[flow] (rebuild.south) -- ([xshift=9mm]decode.north);
\draw[flow] (decode) -- (agg);
\draw[flow] (nocache.south) -- (8.9,-5.65) -- (8.9,-8.8) -- ([yshift=1.5mm]agg.east)
  node[glab, pos=0.32, right=0.5pt, align=left]{out-of-coverage\\ count (not evidence)};
\draw[flow] (agg) -- (result);
\draw[flow] (result) -- (stop);

\node[font=\small\itshape, align=center] at (6.2,-11.85)
  {Optional lanes: hybrid prefilter, verifier/reranker, joint re-read (§4.5, §4.7).};

\end{tikzpicture}
\caption{KV-LTM read flow (per question): each retained session's text is verified against its stored digest first --- a known-good copy is recovered and re-verified on failure, and an unrecoverable session counts as unavailable (out of coverage), never as evidence. Valid state is read from a request-local working copy; when policy allows, invalid or evicted state is rebuilt from the verified text (§4.3), otherwise the session is counted as not cached and delegated to the §4.8 text fallback, outside the depicted cache-read path. Per-session outputs supply bounded evidence to the aggregator (§4.6); optional hybrid, verifier/reranker and joint-evidence lanes (§4.5, §4.7) complement the reference read. Boxes denote logical steps or stored objects, not software components.}
\label{fig:read-flow}
\end{figure}

This chapter specifies the KV-LTM design: what is stored, under which
conditions a stored state is valid, how writes and reads work, how
answers are aggregated, and how the store is retained, invalidated, and
gracefully degraded. Figures~\ref{fig:write-flow}
and~\ref{fig:read-flow} show the write and read flows;
§5--§7 then size it.

\hypertarget{storage-record}{%
\subsection{Storage record}\label{storage-record}}

Each session is stored as one record:

\begin{longtable}[]{@{}
  >{\raggedright\arraybackslash}p{(\columnwidth - 2\tabcolsep) * \real{0.5000}}
  >{\raggedright\arraybackslash}p{(\columnwidth - 2\tabcolsep) * \real{0.5000}}@{}}
\toprule\noalign{}
\begin{minipage}[b]{\linewidth}\raggedright
Field
\end{minipage} & \begin{minipage}[b]{\linewidth}\raggedright
Content
\end{minipage} \\
\midrule\noalign{}
\endhead
\bottomrule\noalign{}
\endlastfoot
\texttt{state} & The session's \textbf{native cache state} under the
reader model: per-token K/V (or MLA latent) for attention layers, plus
the fixed-size recurrent state for hybrid layers (§2.3, §5), plus any
runtime auxiliary state required for replay --- in a versioned state
format \\
\texttt{text} & The raw session text --- the memory of record \\
\texttt{meta} & Timestamps, session ID, model/config hash \\
\texttt{serialization} & Versioned schema: roles, timestamps, prompt
template, tokenizer version, metadata schema version \\
\texttt{text\_digest} / \texttt{state\_digest} & Separate integrity
hashes over the serialized text and the native state. Bad or missing
\textbf{state} is rebuilt only from verified-intact source text; if
\textbf{source} verification fails, the session is recovered from a
known-good copy or marked unavailable/out-of-coverage --- damaged text
is never silently trusted (§4.4, §4.8) \\
\end{longtable}

Serialization is versioned so that a record remains decodable across
software upgrades. We distinguish \textbf{faithful replay} (the entire
§4.2 contract holds and the saved state is intact → the state is
reusable as-is) from \textbf{migration} (a tokenizer change invalidates
the old state but not the recoverability of the retained source text,
which can always be re-prefilled --- §4.8).

\hypertarget{cache-validity-contract}{%
\subsection{Cache-validity contract}\label{cache-validity-contract}}

A stored state is reusable for a read \textbf{when} the read preserves
the §2.2 conditions --- the contract treats them as the operational
checklist (the lemma states sufficiency, not necessity; practical
equality is verified with tolerances by replay, §8): the same complete
token prefix (including fixed instructions), the same model weights,
continued position indices (cache holds 0..N−1, the appended query takes
N..N+Q−1), pinned RoPE base/scaling/max-position for the layers that are
position-parameterized (a hybrid model's MLA layers may be
position-free, §2.2), the same attention masks, the same numerical
configuration, and \textbf{the same native cache state} --- restored
exactly as the writing model computed it, in its native form, with no
format conversion, post-hoc quantization, or pruning between save and
restore (§2.2 boundary i). A matching dtype label is necessary but not
sufficient for equivalent state evolution. The model that
\textbf{writes} the state is the model that \textbf{reads} it; the
PA/answering/aggregator model may differ. (Engineering note, uncited
mechanism remark: some engines show first-token/attention-sink artefacts
when the leading prefix is dropped --- the fixed-prefix contract
prevents this.)

\hypertarget{write-path}{%
\subsection{Write path}\label{write-path}}

At session end, the system persists the session record. The defining
invariant is:

\begin{quote}
\textbf{state = prefill\_reader(fixed\_prefix ‖ serialize(text))} ---
the state is a canonical prefill of the serialized record under the
reader model.
\end{quote}

A live serving cache is reused only when the reader \emph{is} the
serving model and the serialized context is identical; otherwise the
tier prefills the session text (assistant turns included as text tokens)
under the reader model at session end. This one-time ingestion compute
is the amortized term in §7 --- writes add no summarization or
extraction step, but they do add this prefill. Checkpoints are taken
\textbf{per session}: the stored state is the canonical reader prefill
of the complete serialized session at its boundary (assistant turns
included as text tokens). Under ideal arithmetic this equals the live
cache only in the reuse case above. The fixed instruction prefix is
\textbf{shared across sessions}; its bytes are counted once in §5 only
if physically deduplicated, immutable, compatible prefix blocks are
shared --- a hybrid's recurrent boundary state remains per session. A
session the user resumes later can be extended by prefix-exact appending
under the same contract.

\hypertarget{read-path-reference-exhaustive}{%
\subsection{Read path --- reference
(exhaustive)}\label{read-path-reference-exhaustive}}

The reference read path needs \textbf{no retrieval index}. It is the
\emph{fidelity} reference; §7 models its cost and shows when byte
transfer rather than prefill dominates. Per-session outputs are
structured and request-local --- nothing is written back into the
persisted checkpoint:

\begin{enumerate}
\def\labelenumi{\arabic{enumi}.}
\tightlist
\item
  For each retained session: check validity (contract + intact state).
  On a valid hit, restore a \textbf{request-local working copy} of the
  cached prefix.
\item
  Append the question (a short suffix, typically a sentence or two ---
  including the query-time instructions: answer only from this session,
  abstain if unsupported, quote evidence with its date).
\item
  Decode a short per-session answer. Answering and evidence are separate
  fields: a session can abstain from \emph{answering} while still
  supplying \emph{relevant evidence} (a bounded, source-linked partial)
  --- only sessions with no relevant evidence reduce to a count.
\item
  On failed source verification: recover a known-good copy and
  \textbf{re-verify the recovered source}; on success, fall through to
  the state-validity check against the recovered text/serialization
  (load compatible state or rebuild from the verified text, then
  decode); on failure, append an explicit unavailable mark and continue
  --- never rebuild from unverified text (§4.1, §4.8).
\item
  Aggregate the structured outputs (§4.6). Every answer reports its
  coverage: ``state-read n sessions, text-fallback m, abstained a,
  unavailable u'' --- no silent truncation (§8).
\end{enumerate}

The reference pseudocode below permits rebuilding from verified text. A deployment may disable this rebuild path; Figure~\ref{fig:read-flow} then counts the session outside the cache-read coverage and delegates it to the text fallback in §4.8, reported as text-fallback $m$ rather than state-read $n$. This cache-coverage count is not evidence and does not mark the verified text unavailable; any fallback evidence and costs are accounted for separately.

\noindent\begin{minipage}{\linewidth}
\begin{Verbatim}[fontsize=\small]
# prepare: verified source -> request-local state, or None
function prepare(s):
    if not intact(s.text_digest):
        s = recover_known_good(s)
        if s is None or not intact(s.text_digest):
            return None
    if valid(s.meta) and intact(s.state_digest):
        return load_working_copy(s)
    return rebuild_from_text(s)

function read_sessions(q, sessions):
    outs = []
    for s in sessions:
        w = prepare(s)
        if w is None:
            outs.append(unavailable(s))
            continue
        outs.append(decode(w, q))
    return aggregate(q, structured(outs))

function answer_reference(q, store):
    return read_sessions(q, store.retained_sessions())

function answer_verifier(q, candidates):
    return read_sessions(q, candidates)
\end{Verbatim}
\end{minipage}

Here \texttt{prepare} verifies the memory of record before state use;
recovered text is re-verified and retains the original session identity.
State validity includes the full §4.2 contract against that verified
text and serialization. Rebuilds are canonical prefills of verified text
(§4.3); valid state is loaded into a request-local working copy.
\texttt{decode} appends the question and instructions, prefills that
suffix and produces a short output. The reference reads the deduplicated
retained set $R(N)\cup L(M)$ (§4.8); verifier candidates come from an
external retriever (§4.5) and receive the same preparation checks.
\texttt{structured} keeps session IDs, dates, evidence quotes and
answer/abstain fields; unavailable marks are out-of-coverage records,
not evidence, and never enter the aggregator's evidence pool. Answer
abstention does not discard relevant partial evidence (§4.6); coverage
counts are retained separately.

\textbf{Optional query-suffix hints.} The appended question and
query-time instructions may also carry bounded retrieval-derived hints
without changing the stored session prefix. For example, ranked entity
candidates with dated status may help interpret an ambiguous reference.
Hints are clearly delimited, unverified interpretation aids, not
session-local evidence: their rank is not proof, and the reader must
support its output from the session or abstain while retaining relevant
partial evidence. Repeating a hint across session outputs does not
create independent confirmation. These provenance distinctions also
apply at aggregation (§4.6).

A hint producer is outside the unaugmented reference path; any evaluated
hint-augmented variant freezes its source, generation and ranking rules,
suffix format and token cap before testing (§8). Comparisons must
declare and control hint access, report producer and repeated
suffix-processing costs (§7), and preserve the context-fit rule (§4.7,
§8.3). This permits query-side composition with retrieval methods, not a
guarantee of equal or better answer accuracy.

A question supported within one session can be handled by the
per-session reader and bounded aggregator without a joint read.
Cross-session partials may also suffice for aggregation; otherwise the
declared §4.7 trigger may invoke a bounded joint-text read or follow-up.
This dispatch is a protocol choice, not a claim that every single-fact
question succeeds or that every comparison requires joint attention.

\hypertarget{read-path-optional-hybrid-variant-and-the-verifierreranker-role}{%
\subsection{Read path --- optional hybrid variant and the
verifier/reranker
role}\label{read-path-optional-hybrid-variant-and-the-verifierreranker-role}}

\textbf{Hybrid variant.} A \textbf{pre-load prefilter} over
\emph{text/metadata} (metadata filters, lexical match, or a dense
embedding variant) selects candidate sessions before any state is
loaded. Only the prefilter avoids cold-cache I/O: a query-conditioned
1-token probe still loads and reads each candidate cache --- it saves
decode, not cold reads. The prefilter method and any embedder are
\textbf{frozen protocol parameters, fixed before evaluation runs}; a
dense variant's index is not assumed small; selection effects are
measured and reported (§8). The exhaustive path remains the reference;
the hybrid is an optional, separately reported variant.

\textbf{Verifier/reranker role.} KV-LTM can also sit \emph{behind} an
existing retrieval system: the retriever ranks candidate sessions, and
KV-LTM reads their cached state --- verifying and reranking candidates
against full session content instead of re-prefilling their text. The
role applies to candidates with valid state; other candidates fall back
to text prefill. This is complementarity mechanism (b) of §1.3;
mechanism (a) is the index-free reference tier above.

\hypertarget{aggregation-contract}{%
\subsection{Aggregation contract}\label{aggregation-contract}}

The aggregator is part of the method, specified enough to be evaluated:

\begin{itemize}
\tightlist
\item
  \textbf{Inputs:} the question; dated, source-linked per-session
  evidence; abstain flags (abstention triggers are protocol parameters,
  §8). Answer abstention is not evidence absence: sessions that abstain
  from answering keep their bounded partial evidence and session IDs in
  the pool; only no-relevant-evidence records reduce to counts.
\item
  \textbf{Bound (explicit and declared):} at most \textbf{S} source
  evidences, each capped at \textbf{e} tokens, output ≤ \textbf{o}
  tokens; abstain flags are not truncated. Overflow follows a declared
  rule --- staged/tree aggregation or a documented ordering --- frozen
  across arms A, A\_k, O and reported. S, e, o are protocol parameters
  fixed before runs (§8); their storage/latency consequences are
  computed in §7.
\item
  \textbf{Policy:} a bounded, source- and recency-aware synthesis over
  dated evidence. \textbf{Majority voting is not truth resolution}:
  repeated stale facts cannot override a supported newer fact by
  frequency alone.
\item
  \textbf{Outputs:} a supported answer with source links \textbar{}
  abstain \textbar{} follow-up (including original-text fallback).
\item
  \textbf{Protocol:} the aggregator's model, prompt, and budget are
  fixed across arms A, A\_k, O (§8).
\end{itemize}

\hypertarget{cross-session-questions}{%
\subsection{Cross-session questions}\label{cross-session-questions}}

Per-session reads cannot jointly attend across sessions. For
joint-evidence questions, a second tier selects the top-k\_joint
relevant sessions --- drawn from the sessions whose per-session outputs
carried relevant evidence, \textbf{including partial evidence from
sessions that abstained from answering} --- and re-prefills them from
text for a joint read. (A bounded reselection over source text under a
frozen rule is allowed when per-session evidence is insufficient.) The
tier is invoked on the aggregator's follow-up output or by question
type; the joint read must fit the reader's context including
instructions, query, and output reserve --- otherwise the system
abstains or follows up (§8 reports joint-fallback limits separately; arm
C, full-concat, is a different system used for contrast, not this tier).
Concatenating independently cached sessions into one cache is non-exact
--- it misses cross-session attention \cite[qualitative]{K6}; naive
concatenation also breaks position indices (our own technical
observation) --- so the joint path works from text, not from fused
caches. (Optimization note: an already-stored first session's cache may
prefix the joint input \textbf{only when that complete saved prefix
matches the actual joint serialization under the entire §4.2 contract}
--- joint template, leading instructions, positions, masks; otherwise
the whole joint input is prefilled.)

\hypertarget{retention-invalidation-fallback}{%
\subsection{Retention, invalidation,
fallback}\label{retention-invalidation-fallback}}

\begin{itemize}
\tightlist
\item
  \textbf{Retention.} The store holds the recent set R(N months) ∪ the
  older set L(M). The capacity invariant applies to the
  \textbf{deduplicated union}, including native-state and metadata
  overhead: bytes(R ∪ L) plus per-query modelled latency must fit the
  budget derived in §6/§7. Under exhaustive reference reads,
  recency-of-load is vacuous (every retained session is read every
  query), so \textbf{``use'' is defined as evidence use}: a session is
  used when it contributed supported evidence the aggregator cited ---
  including cited partial evidence from answer-abstaining sessions ---
  or scored a verifier/reranker hit; L(M) is ordered by last-used under
  this definition, which is a frozen protocol parameter. When either the
  byte or the latency budget is exceeded, the older-set allowance
  shrinks first and the recent window follows --- ``N months'' becomes
  \emph{achievable-N}; overflow sessions drop to text-only. KV eviction
  is not text deletion: evicted sessions remain readable from their
  text.
\item
  \textbf{Invalidation.} A model, tokenizer, config,
  fixed-prefix/prompt-template, or serialization-version change
  invalidates the stored state; affected sessions are re-prefilled from
  text in the background (migration cost quantified in §7).
\item
  \textbf{Fallback.} Evicted or invalid state still leaves the original
  text: full-prefill reading, or RAG over it when full reading is
  infeasible. Degradation preserves \textbf{availability}; the fidelity
  property holds only on paths that read the full session --- the RAG
  branch inherits §3.1's query-time selection exposure.
\end{itemize}

\hypertarget{d1-logical-flow-and-stored-objects}{%
\subsection{D1 --- logical flow and stored
objects}
Figure~\ref{fig:read-flow} shows the read path's logical steps and their data relationships.\label{d1-logical-flow-and-stored-objects}}

\hypertarget{chapter-5-kv-cache-size-estimation}{%
\section{KV-Cache Size
Estimation}\label{chapter-5-kv-cache-size-estimation}}

This chapter sizes the per-session state that §4 stores. All quantities
are \textbf{bytes per token}; a sequence-length factor is applied only
when totals are computed (Appendix A shows every derivation). GB/TB are
decimal (10⁹/10¹² B) throughout.

\hypertarget{formulas}{%
\subsection{Formulas}\label{formulas}}

\begin{itemize}
\tightlist
\item
  \textbf{MHA / GQA:}
  \texttt{bytes/token\ =\ 2\ ·\ L\ ·\ H\_kv\ ·\ d\_head\ ·\ b} (2 = K+V;
  L = layers; H\_kv = KV heads; d\_head = head dim; b = bytes per
  element).
\item
  \textbf{MLA:} \texttt{bytes/token\ =\ L\ ·\ (d\_c\ +\ d\_R)\ ·\ b}
  (d\_c = KV LoRA rank; d\_R = rope head dim). MLA figures assume a
  runtime that persists the \textbf{compressed latent} (d\_c + d\_R); a
  runtime that caches decompressed per-head K/V stores ≈ H·(d\_qk +
  d\_v)/(d\_c + d\_R)× more (≈8.9× for DeepSeek-V2-Lite; derivation in
  A.1).
\item
  \textbf{Hybrid (KDA + MLA):}
  \texttt{total\ =\ sessions\ ×\ S\_recurrent\ +\ tokens\ ×\ b\_MLA} ---
  the recurrent state is fixed-size per session (symbolic S\_recurrent
  pending runtime/dtype/auxiliary validation); the MLA layers keep
  per-token state.
\end{itemize}

\hypertarget{t1-pinned-config-table}{%
\subsection{T1 --- pinned-config
table}\label{t1-pinned-config-table}}

\begin{longtable}[]{@{}
  >{\raggedright\arraybackslash}p{(\columnwidth - 8\tabcolsep) * \real{0.2000}}
  >{\raggedright\arraybackslash}p{(\columnwidth - 8\tabcolsep) * \real{0.2000}}
  >{\raggedright\arraybackslash}p{(\columnwidth - 8\tabcolsep) * \real{0.2000}}
  >{\raggedright\arraybackslash}p{(\columnwidth - 8\tabcolsep) * \real{0.2000}}
  >{\raggedright\arraybackslash}p{(\columnwidth - 8\tabcolsep) * \real{0.2000}}@{}}
\toprule\noalign{}
\begin{minipage}[b]{\linewidth}\raggedright
Model (pinned config)
\end{minipage} & \begin{minipage}[b]{\linewidth}\raggedright
Class
\end{minipage} & \begin{minipage}[b]{\linewidth}\raggedright
Bytes/token (bf16)
\end{minipage} & \begin{minipage}[b]{\linewidth}\raggedright
TB per 1M tokens
\end{minipage} & \begin{minipage}[b]{\linewidth}\raggedright
Annual @30K tok/day*
\end{minipage} \\
\midrule\noalign{}
\endhead
\bottomrule\noalign{}
\endlastfoot
MiniCPM3-4B (62 layers; d\_c 256 + d\_R 32) & MLA & 35,712 & 0.036 &
≈0.39 TB \\
DeepSeek-V2-Lite (27 layers; d\_c 512 + d\_R 64; MoE 16B/2.4B active) &
MLA & 31,104 & 0.031 & ≈0.34 TB \\
Qwen2.5-7B (28 layers × 4 kv-heads × 128) & GQA & 57,344 & 0.057 & ≈0.63
TB \\
Llama-3.1-8B (32 layers × 8 × 128) & GQA & 131,072 & 0.131 & ≈1.44 TB \\
Llama-3.3-70B (80 layers × 8 × 128) & GQA & 327,680 & 0.328 & ≈3.59
TB \\
DeepSeek-V3-class (61 layers × 576) --- \textbf{published-config class
assumption, not a pinned checkpoint} & MLA & 70,272 & 0.070 & ≈0.77 TB
bf16 / ≈0.39 TB ideal FP8** \\
Kimi-Linear-48B-A3B (27 layers = 20 KDA + 7 MLA × 576) & hybrid (KDA +
MLA) & b\_MLA = 8,064 & 0.008 + (500 × S\_recurrent)/10¹² & ≈0.088 TB
(MLA part) + 5,475 × S\_recurrent B \\
\end{longtable}

* The 30K tokens/day scenario is a declared assumption (§6). ** Ideal
FP8 halves the \emph{quantized payload} only (before scales/metadata;
recurrent/non-quantized state unaffected) and is numerically lossy
(class ii, §2.3) --- validate per config before relying on it \cite{K12}.
Config values from public HF config.json, retrieved 2026-09-26
\cite{K11}: MiniCPM3-4B, DeepSeek-V2-Lite, Qwen2.5-7B, Llama-3.1-8B,
Llama-3.3-70B; Kimi-Linear decomposition per \cite{K15} plus its public
config. The DeepSeek-V3-class row uses the published architecture's
parameters as a \textbf{representative class assumption} --- labelled,
not a pinned checkpoint. Context caps: 32,768 max positions for
MiniCPM3-4B (config value); \textasciitilde32K for DeepSeek-V2-Lite
(model-card guidance; its config allows more under scaling --- config ≠
effective context, §9). Sessions longer than the cap need segmentation,
noted in §9. The shared fixed instruction prefix (§4.3) is stored once
as a physically deduplicated, immutable prefix block; its contribution
is negligible at the declared volumes (a 500-token prefix ≈ 65.5 MB once
≈ 0.005\% of the 8B-GQA annual state) --- scenario token volumes count
session tokens only.

\hypertarget{the-storage-contrast}{%
\subsection{The storage contrast}\label{the-storage-contrast}}

KV state is large relative to text: for the 8B-GQA config, 131,072
B/token vs ≈4.5 B/token of raw text (6 B/word ÷ 1.33 tokens/word,
declared planning ratios) --- a ratio of ≈2.9×10⁴ (derived per config in
Appendix A; no universal multiplier is claimed). KV-LTM increases
storage; what it buys is avoided re-prefill compute and the absence of
write-time derivation loss (§3). The decision variable is therefore not
``is the cache small'' but ``does the retention budget (§6) keep the
working set affordable per query (§7)''.

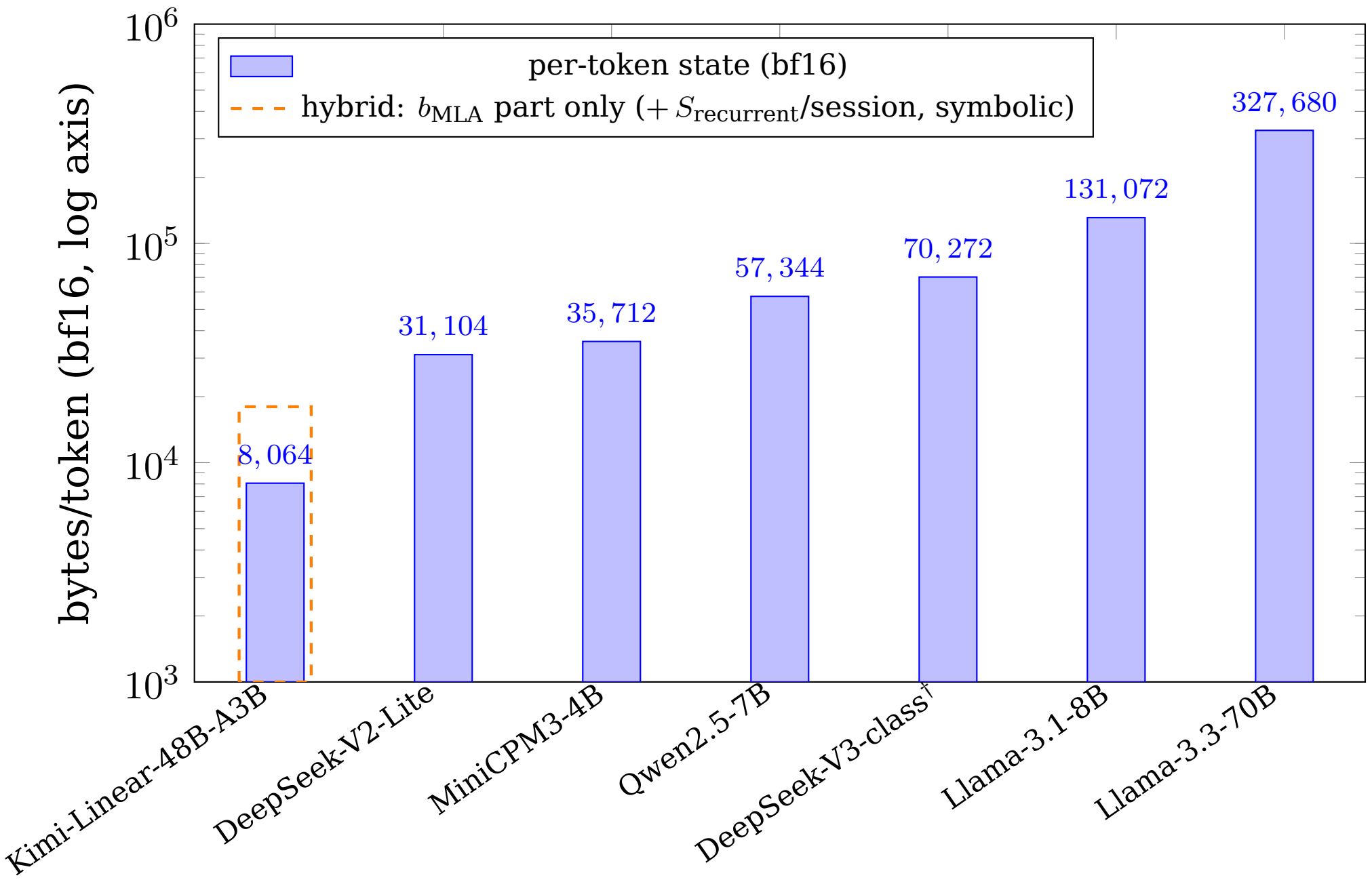
\begin{figure}[t]
\centering
\begin{tikzpicture}
\begin{axis}[
  ylabel={bytes/token (bf16, log axis)},
  ymode=log, log origin=infty,
  ymin=1e3, ymax=1e6,
  ytick={1e3,1e4,1e5,1e6},
  yticklabels={$10^3$,$10^4$,$10^5$,$10^6$},
  symbolic x coords={Kimi-Linear-48B-A3B, DeepSeek-V2-Lite, MiniCPM3-4B, Qwen2.5-7B, DeepSeek-V3-class, Llama-3.1-8B, Llama-3.3-70B},
  xtick={Kimi-Linear-48B-A3B, DeepSeek-V2-Lite, MiniCPM3-4B, Qwen2.5-7B, DeepSeek-V3-class, Llama-3.1-8B, Llama-3.3-70B},
  xticklabels={Kimi-Linear-48B-A3B, DeepSeek-V2-Lite, MiniCPM3-4B, Qwen2.5-7B, DeepSeek-V3-class$^\dagger$, Llama-3.1-8B, Llama-3.3-70B},
  enlarge x limits=0.08,
  x tick label style={rotate=35, anchor=east, font=\scriptsize},
  nodes near coords={\pgfmathprintnumber[fixed, precision=0, 1000 sep={,}]{\pgfplotspointmeta}},
  point meta=rawy,
  nodes near coords style={font=\scriptsize},
  width=13cm, height=8cm,
  legend style={at={(0.02,0.98)}, anchor=north west, font=\scriptsize}
]
\addplot+[ybar, bar width=16pt, fill=blue!25, mark=none, area legend] coordinates {
  (Kimi-Linear-48B-A3B, 8064)
  (DeepSeek-V2-Lite, 31104)
  (MiniCPM3-4B, 35712)
  (Qwen2.5-7B, 57344)
  (DeepSeek-V3-class, 70272)
  (Llama-3.1-8B, 131072)
  (Llama-3.3-70B, 327680)
};
\addlegendentry{per-token state (bf16)}
\draw[orange, dashed, thick] ([xshift=-10pt] axis cs:Kimi-Linear-48B-A3B,1e3) rectangle ([xshift=10pt] axis cs:Kimi-Linear-48B-A3B,1.8e4);
\addlegendimage{orange, dashed, thick}
\addlegendentry{hybrid: $b_{\mathrm{MLA}}$ part only ($+\,S_{\mathrm{recurrent}}$/session, symbolic)}
\end{axis}
\end{tikzpicture}
\caption{BF16 state bytes per token, logarithmic scale, for six pinned configurations and one representative DeepSeek-V3-class assumption (marked \dag, not a pinned checkpoint). MLA values assume compressed-latent storage. Kimi-Linear shows only its MLA-layer component (b\textsubscript{MLA} = 8,064 B/token); its per-session recurrent state is symbolic and not drawn. The plotted per-token spread is $\sim$40\texttimes{} (the hybrid's b\textsubscript{MLA} part at the low end); among the non-hybrid configs the spread is ≈10.5\texttimes{} (31,104 \textrightarrow{} 327,680).}
\label{fig:c1}
\end{figure}

\hypertarget{c1-bytestoken-across-the-pinned-configs}{%
\subsection{C1 --- bytes/token across the pinned
configs}
Figure~\ref{fig:c1} shows the per-token spread.\label{c1-bytestoken-across-the-pinned-configs}}

\hypertarget{chapter-6-storage-estimation-for-the-pa-use-case}{%
\section{Storage Estimation for the PA Use
Case}\label{chapter-6-storage-estimation-for-the-pa-use-case}}

This chapter sizes the store of §4 for one user's personal assistant,
per year. Every volume is a \textbf{declared, labelled assumption} ---
not a behavioral fact; public literature gives order-of-magnitude
context only \cite{K35}. Appendix A shows the arithmetic, and Appendix
A.2 states the declared assumptions behind each volume.

\hypertarget{volume-scenarios}{%
\subsection{Volume scenarios}\label{volume-scenarios}}

One coverage parameter drives everything: the fraction of a user's daily
language activity that flows through the PA.

\begin{longtable}[]{@{}llll@{}}
\toprule\noalign{}
Scenario & Coverage (declared assumption) & Tokens/day & Tokens/year \\
\midrule\noalign{}
\endhead
\bottomrule\noalign{}
\endlastfoot
Low & 12\% & ≈10,000 & 3.65M \\
Mid (anchor) & 35\% & ≈30,000 & 10.95M \\
High & 100\% & ≈85,500 & 31.2M \\
\end{longtable}

Context anchors for plausibility: measured speech production is ≈16K
spoken words/day per adult \cite{K35} --- order-of-magnitude context
only; the calibration below uses a ≈13K-word speech component from its
own recorded sources. The full-stream calibration behind the high row
(speech + hearing + reading + writing ≈ 64K words/day), the ≈1.33
tokens/word and ≈4.5 B/token (6 B/word) ratios, and the coverage
fractions (35\% is a mid-point planning anchor, not a measurement) are
declared assumptions (Appendix A.2). Tokens/word remains a parameter to
measure on the chosen tokenizer \cite{K36}.

\hypertarget{t2-annual-storage-per-scenario}{%
\subsection{T2 --- annual storage per
scenario}\label{t2-annual-storage-per-scenario}}

Bytes from the pinned §5 table; totals in Appendix A. The text-only
column keeps the storage contrast: KV-LTM stores \emph{more} than text;
the benefit is avoided re-prefill compute and no write-time derivation
loss (§3, §5.3).

\begin{longtable}[]{@{}
  >{\raggedright\arraybackslash}p{(\columnwidth - 8\tabcolsep) * \real{0.2000}}
  >{\raggedright\arraybackslash}p{(\columnwidth - 8\tabcolsep) * \real{0.2000}}
  >{\raggedright\arraybackslash}p{(\columnwidth - 8\tabcolsep) * \real{0.2000}}
  >{\raggedright\arraybackslash}p{(\columnwidth - 8\tabcolsep) * \real{0.2000}}
  >{\raggedright\arraybackslash}p{(\columnwidth - 8\tabcolsep) * \real{0.2000}}@{}}
\toprule\noalign{}
\begin{minipage}[b]{\linewidth}\raggedright
Model (class)
\end{minipage} & \begin{minipage}[b]{\linewidth}\raggedright
Low (3.65M tok)
\end{minipage} & \begin{minipage}[b]{\linewidth}\raggedright
Mid (10.95M tok)
\end{minipage} & \begin{minipage}[b]{\linewidth}\raggedright
High (31.2M tok)
\end{minipage} & \begin{minipage}[b]{\linewidth}\raggedright
Sensitivity
\end{minipage} \\
\midrule\noalign{}
\endhead
\bottomrule\noalign{}
\endlastfoot
Llama-3.1-8B (GQA) & 0.48 TB & 1.44 TB & 4.09 TB & linear in coverage;
achievable-N at a fixed budget scales \textbf{inversely} with
coverage \\
Llama-3.3-70B (GQA) & 1.20 TB & 3.59 TB & 10.2 TB & same linearity \\
Qwen2.5-7B (GQA) & 0.21 TB & 0.63 TB & 1.79 TB & \\
DeepSeek-V3-class (MLA bf16) & 0.26 TB & 0.77 TB & 2.19 TB & ideal-FP8
halves the quantized payload only (before scales/metadata; lossy ---
validate) \\
DeepSeek-V2-Lite (MLA) & 0.11 TB & 0.34 TB & 0.97 TB & \\
MiniCPM3-4B (MLA) & 0.13 TB & 0.39 TB & 1.11 TB & \\
Kimi-Linear-48B-A3B (hybrid) & 0.03 TB + 1,825×S\_rec B & 0.088 TB +
5,475×S\_rec B & 0.25 TB + 15,600×S\_rec B & S\_recurrent
(bytes/session) symbolic until validated; session counts from the
declared 2,000-token sessions \\
\textbf{Text-only contrast} & \textbf{≈16 MB raw (≈5.5 MB at 3:1)} &
\textbf{≈49 MB raw (≈16 MB at 3:1)} & \textbf{≈141 MB raw (≈47 MB at
3:1)} & 3:1 is an illustrative workload-dependent planning assumption,
inside the 2.9--3.6× range documented for selected settings on the
mixed-data Silesia corpus \cite{K37} --- not validated PA compression \\
\end{longtable}

\hypertarget{retention-set-size-from-a-byte-budget}{%
\subsection{Retention set size from a byte
budget}\label{retention-set-size-from-a-byte-budget}}

§4's retention --- R(N months) ∪ L(M), capacity from a budget ---
becomes concrete once a store budget is declared. Example with a
declared B\_store = 512 GB (a single consumer-NVMe allocation;
assumption, not a measurement), mid scenario, uniform session sizes
(2,000 tokens, declared), \textbf{allocating the whole budget to the
recent set (L = 0)}, token payload only --- the table divides by state
bytes; the §4.8 invariant over the full record (incl.~overhead) is
checked in the §8 replay. This is a \textbf{byte-only upper estimate}:

\begin{longtable}[]{@{}
  >{\raggedright\arraybackslash}p{(\columnwidth - 4\tabcolsep) * \real{0.3333}}
  >{\raggedright\arraybackslash}p{(\columnwidth - 4\tabcolsep) * \real{0.3333}}
  >{\raggedright\arraybackslash}p{(\columnwidth - 4\tabcolsep) * \real{0.3333}}@{}}
\toprule\noalign{}
\begin{minipage}[b]{\linewidth}\raggedright
Model
\end{minipage} & \begin{minipage}[b]{\linewidth}\raggedright
TB/yr (mid)
\end{minipage} & \begin{minipage}[b]{\linewidth}\raggedright
Achievable-N within 512 GB (upper estimate)
\end{minipage} \\
\midrule\noalign{}
\endhead
\bottomrule\noalign{}
\endlastfoot
Llama-3.1-8B & 1.44 & ≈4.3 months \\
DeepSeek-V3-class (bf16) & 0.77 & ≈8.0 months \\
MiniCPM3-4B & 0.39 & ≈16 months \\
DeepSeek-V2-Lite & 0.34 & ≈18 months \\
Kimi-Linear (hybrid) & 0.088 + sessions×S\_rec & formula below; 5.8
years is the optimistic \textbf{MLA-only byte ceiling} (S\_rec → 0), not
attainable full-hybrid retention \\
\end{longtable}

For the hybrid,
\texttt{N\ ≤\ B\_store\ /\ (tokens/yr\ ×\ b\_MLA\ +\ sessions/yr\ ×\ S\_recurrent\ +\ overhead/yr)}
(all terms in bytes). Illustrative sensitivity: at S\_recurrent ≈ 21
MB/session and ≈5,475 sessions/yr (mid), the recurrent share ≈115 GB/yr
exceeds the 88 GB/yr MLA part, giving N ≈ 512e9 / (88.3e9 + 115e9) ≈ 2.5
years (derivation and its limits in A.5).

Reading: among the fully specified configs the model choice moves
achievable-N by ≈4× (4.3 → 18 months) --- and more if the hybrid's
recurrent state proves small. The budget, not a calendar, decides the
retention window; when it is exceeded, the oldest sessions drop to
text-only (achievable-N, §4.8) --- the store degrades instead of
failing. We report TB, not \$/TB: storage prices are dated and
deployment-specific; a dated price source can be added at deployment
time without changing the model.

\hypertarget{what-this-chapter-does-not-claim-ch6}{%
\subsection{What this chapter does not
claim}\label{what-this-chapter-does-not-claim-ch6}}

The volumes are planning scenarios with declared coverage parameters,
not measured PA usage. The byte figures follow from pinned model configs
(Appendix A). No storage-saving claim is made for KV-LTM --- it strictly
increases storage over text-only memory; the trade it offers is compute
and write-time fidelity (§3, §7).
\hypertarget{chapter-7-compute-estimation-enabling-techniques}{%
\section{Compute Estimation \& Enabling
Techniques}\label{chapter-7-compute-estimation-enabling-techniques}}

This chapter models the per-query cost of the §4 read paths. No single
cost dominates universally --- the regime decides. All latency figures
are \textbf{analytic estimates over declared parameters} until the §8
minimal real-cache replay measures them; nothing here is a measured
result.

\hypertarget{per-query-cost-model}{%
\subsection{Per-query cost model}\label{per-query-cost-model}}

\noindent\begin{minipage}{\linewidth}
For a query over the retained set $C = R(N) \cup L(M)$ (§4.8):
\begin{Verbatim}[fontsize=\small]
T(q) ≈ B_cold(q) / β_eff              (a)
     + T_attn(q)                      (b)
     + Σ_s g_s / γ                    (c)
     + T_agg(S, e, o)                 (d)
     + Σ_{s∈rebuild} T_prefill(n_s)   (e)
     + T_joint(k_joint)               (f)
     + T_prefilter                    (g)
     + T_model                        (h)
     − overlap                        (i)
\end{Verbatim}
\small
\begin{itemize}
\setlength{\itemsep}{2pt}
\setlength{\parskip}{0pt}
\item[(a)] \textbf{Cold transfer.} Bytes moved over effective bandwidth;
$\beta_{\mathrm{eff}}$ is symbolic or based on a named, dated source,
distinguishing theoretical datasheet bandwidth from achieved transfer.
\item[(b)] \textbf{Query-prefill attention.} Attention over the cached
lengths when processing the appended suffix.
\item[(c)] \textbf{Decode work.} $g_s$ is each session's output-token
count ($C\times g$ in total when equal). Declare one convention for
$\gamma$ per run: serial aggregate (one active stream) or ideal parallel
aggregate (sum over streams). Assumed rates are not achieved latencies.
Throughput depends on bandwidth, residency and compute; dense attention
re-reads cached state per generated token. Thus $\gamma$ is not an
independent scaling knob; do not double-count attention or model work
already included in measured $\gamma$.
\item[(d)] \textbf{Bounded aggregation.} Uses $S,e,o$ (§4.6); staged or
tree aggregation counts every call.
\item[(e)] \textbf{Conditional rebuild.} Query-time prefill of invalid
or missing state (§4.4), distinct from background migration; zero when
all state is valid.
\item[(f)] \textbf{Conditional joint tier.} Joint-tier text re-prefill
(§4.7); zero when that tier is not invoked.
\item[(g)] \textbf{Hybrid prefilter.} The $A_k$ prefilter (§4.5); zero
on the reference path.
\item[(h)] \textbf{Residual model compute.} Model work not covered above.
\item[(i)] \textbf{Overlap correction.} Subtract only overlap justified
by a measured schedule from the otherwise non-overlapped sum.
\end{itemize}
\end{minipage}

plus the amortized ingestion prefill from §4.3 (one-time per session)
and the re-prefill on model change (§4.8) --- linear in archive
\emph{token volume} for fixed/bounded session-length processing
(full-attention FLOP scaling is a separate term we do not expand here).
Note the \textbf{coupling}: the decode term is what forces each
session's state to be read --- bytes and decode are not independent
alternatives.

\hypertarget{worked-examples-declared-assumptions}{%
\subsection{Worked examples (declared
assumptions)}\label{worked-examples-declared-assumptions}}

Assume: 2,000-token sessions, g = 64 output tokens/session, every
selected state cold (no residency) --- all declared assumptions, not
measurements.

\begin{itemize}
\item
  \textbf{Recent window, 8B-GQA (131,072 B/token):} 300 sessions ≈ 600K
  tokens ≈ 79 GB of state --- about 20 days at the §6 mid scenario. Per
  exhaustive query: cold transfer ≈ 79 GB/β\_eff; decode ≈ 300 × 64 ≈
  19K generated tokens.
\item
  \textbf{The coupling, made concrete:} at a declared β\_eff = 5 GB/s
  the 79 GB cold transfer takes ≈15.7 s. Decode of 19,200 tokens: under
  a declared serial aggregate γ = 30 tokens/s, ≈640 s of serial decode;
  under an ideal 300-way parallel setup sustaining 30 tokens/s on every
  stream (aggregate γ = 9,000 --- requiring additional resources), ≈2.1
  s elapsed. Both γ figures are assumptions, not achieved latencies ---
  and γ is not independent of transfer: dense attention re-reads the
  cached state each generated token (≈ g × B\_state ≈ 5 TB of logical KV
  reads for this example), so actual device traffic and overlap are
  runtime-dependent. Which term dominates is regime-dependent; §8
  reports both terms per arm, measured.
\item
  \textbf{Year-scale, same model:} 10.95M tokens ≈ 1.44 TB of state per
  exhaustive query; the cold-transfer component is 1.44 TB / β\_eff (at
  the same declared 5 GB/s, ≈287 s) --- note this is the transfer
  component of the query, not the full-query duration. This is the
  practical motivation for the §4.5 prefilter: the hybrid variant
  evaluates how much of that scan is avoidable, with its selection
  effects measured (§8).
\item
  \textbf{Same volumes under MLA/hybrid configs} shrink by the §5 ratios
  for the validated per-token parts (e.g., DeepSeek-V2-Lite: 0.34 TB/yr,
  ≈4× smaller than 8B-GQA); unvalidated recurrent/auxiliary state is
  excluded from these ratios until measured.
\item
  \textbf{Concurrent residency on one GPU (declared assumptions).}
  Independent session reads can be batched, with each read using its own
  request-local state. Table T3 estimates memory capacity, not achieved
  concurrency or latency: roughly 385--850 8B-GQA reads or 1,400--3,300
  reads for the two MLA configurations fit under these assumptions.
  Longer sessions reduce the counts. The assumed 70B weight allocation
  does not fit within the reserved-memory budget of one H200; it leaves
  space for roughly 30 reads on B200 or 165 on B300. Cold transfer,
  decode compute and queueing remain separate constraints (§7.1); actual
  performance must be measured (§8).
\end{itemize}

\textbf{T3 --- session states resident on one GPU (analytic estimate,
not measured).} NVIDIA-listed HBM capacities \cite{K57}; BF16 state per
§5; 2,000 session tokens plus a 128-token appended question/instruction
suffix and 64 output tokens; one identical 500-token fixed prefix
physically shared across reads; 10\% of HBM reserved for runtime needs.
BF16 model-weight allocations are illustrative budgets, with all experts
resident for MoE. Counts are rounded; A.6 gives the formula and integer
results conditional on these assumptions.

\begin{longtable}[]{@{}
  >{\raggedright\arraybackslash}p{(\columnwidth - 10\tabcolsep) * \real{0.3293}}
  >{\raggedright\arraybackslash}p{(\columnwidth - 10\tabcolsep) * \real{0.1220}}
  >{\raggedright\arraybackslash}p{(\columnwidth - 10\tabcolsep) * \real{0.1463}}
  >{\raggedright\arraybackslash}p{(\columnwidth - 10\tabcolsep) * \real{0.1341}}
  >{\raggedright\arraybackslash}p{(\columnwidth - 10\tabcolsep) * \real{0.1341}}
  >{\raggedright\arraybackslash}p{(\columnwidth - 10\tabcolsep) * \real{0.1341}}@{}}
\toprule\noalign{}
\begin{minipage}[b]{\linewidth}\raggedright
Config (BF16)
\end{minipage} & \begin{minipage}[b]{\linewidth}\raggedright
Weight budget
\end{minipage} & \begin{minipage}[b]{\linewidth}\raggedright
State per read
\end{minipage} & \begin{minipage}[b]{\linewidth}\raggedright
H200 (141 GB)
\end{minipage} & \begin{minipage}[b]{\linewidth}\raggedright
B200 (180 GB)
\end{minipage} & \begin{minipage}[b]{\linewidth}\raggedright
B300 (288 GB)
\end{minipage} \\
\midrule\noalign{}
\endhead
\bottomrule\noalign{}
\endlastfoot
MiniCPM3-4B (MLA) & 8.15 GB & ≈78 MB & ≈1,500 & ≈2,000 & ≈3,200 \\
DeepSeek-V2-Lite (MLA, MoE; all experts resident) & 31.4 GB & ≈68 MB &
≈1,400 & ≈1,900 & ≈3,300 \\
Qwen2.5-7B (GQA) & 15.2 GB & ≈126 MB & ≈890 & ≈1,170 & ≈1,940 \\
Llama-3.1-8B (GQA) & 16.1 GB & ≈287 MB & ≈385 & ≈510 & ≈850 \\
Llama-3.3-70B (GQA) & 141.1 GB & ≈718 MB & 0 (no fit) & ≈30 & ≈165 \\
\end{longtable}

\hypertarget{enabling-techniques}{%
\subsection{Enabling techniques}\label{enabling-techniques}}

\begin{itemize}
\tightlist
\item
  \textbf{Tiered KV store} (HBM/DRAM/NVMe): cross-query KV reuse and
  tiered KV storage \cite{K26}; disaggregated CPU/DRAM/SSD designs
  \cite{K27}; SSD-backed KV serving \cite{K29} --- all qualitative;
  tested-setup magnitudes are not claimed here.
\item
  \textbf{Prefix-cache infrastructure}: PagedAttention/vLLM \cite{K23}
  and RadixAttention/SGLang \cite{K24} are the in-memory substrate our
  persistence extends.
\item
  \textbf{KV quantization} (FP8/INT4): reduces the quantized payload
  (≈2×/4× ideal, before scales/metadata; weights and non-quantized
  recurrent state unaffected) but is numerically lossy (class ii, §2.3)
  --- validate per config before counting it \cite{K12}.
\item
  \textbf{Batching and session-level parallelism}: per-session reads are
  independent, so fan-out parallelizes across devices/processes; the
  aggregate work is unchanged (parallelism trades latency for resources;
  a single-GPU residency estimate is given in §7.2, T3).
\end{itemize}

\hypertarget{what-this-chapter-does-not-claim-ch7}{%
\subsection{What this chapter does not
claim}\label{what-this-chapter-does-not-claim-ch7}}

No FLOPs-derived latency is presented as a real restore latency ---
those are marked \emph{to-measure} via the §8 replay. Bandwidth values
are symbolic or declared assumptions, not datasheet facts. And the model
deliberately reports regimes: which term dominates depends on retention
size, residency, and hardware --- that dependence is the point of §8's
measured cost axis.
\hypertarget{chapter-8-evaluation-design-testing-without-implementation}{%
\section{Evaluation Design: Testing Without
Implementation}\label{chapter-8-evaluation-design-testing-without-implementation}}

This chapter defines how KV-LTM's logical read path can be measured on
public long-term-memory benchmarks \textbf{without implementing
persistent caching} --- per-session prefill stands in for state
restoration (§2.2's lemma), so the read path is emulated exactly under
ideal arithmetic, with practical equality tested later by a small
real-cache replay.

\textbf{Framing caution.} No arm below is a guaranteed bound on another.
Published results already show a retrieval system outscoring
full-context reading on LongMemEval-S (Zep 71.2\% vs full-context gpt-4o
60.2\% \cite[vendor preprint]{K3}) --- more context is not automatically
better. We therefore compare \textbf{defined systems}, not claimed
ceilings. The Zep figures are vendor-reported under that study's
evaluation protocol, not a controlled KV-LTM experiment. Evaluation
labels and judges can change reported scores (§9.1); the comparison is a
framing caution, not a bound on any arm.

\hypertarget{arms}{%
\subsection{Arms}\label{arms}}

All arms read the \textbf{same retained archive} under the same
retention policy (§4.8), use the §4.1 canonical serialization (roles,
session timestamps), and receive the benchmark's question date; any
whole-original-history variant is a separate, labelled contrast (C
concatenates the common retained history). The aggregator (model,
prompt, bound S/e/o, overflow rule) and the joint tier (k\_joint,
trigger, context cap) are \textbf{frozen before test runs and identical
across A, A\_k, O}.

\begin{longtable}[]{@{}
  >{\raggedright\arraybackslash}p{(\columnwidth - 4\tabcolsep) * \real{0.3333}}
  >{\raggedright\arraybackslash}p{(\columnwidth - 4\tabcolsep) * \real{0.3333}}
  >{\raggedright\arraybackslash}p{(\columnwidth - 4\tabcolsep) * \real{0.3333}}@{}}
\toprule\noalign{}
\begin{minipage}[b]{\linewidth}\raggedright
Arm
\end{minipage} & \begin{minipage}[b]{\linewidth}\raggedright
Definition
\end{minipage} & \begin{minipage}[b]{\linewidth}\raggedright
Role
\end{minipage} \\
\midrule\noalign{}
\endhead
\bottomrule\noalign{}
\endlastfoot
\textbf{A} & Per-session prefill + appended question over ALL retained
sessions → per-session outputs → aggregator (§4.6) + joint tier (§4.7) &
The KV-LTM logical read path, emulated without an implementation \\
\textbf{A\_k\^{}pre / A\_k\^{}probe} & The frozen hybrid variants
(§4.5): pre-load prefilter → top-k → read, resp. 1-token probe → top-k →
read & Optional selected variants, \textbf{reported as separate cost
rows} --- the prefilter saves cold I/O, the probe saves decode; fusing
them would conflate two mechanisms \\
\textbf{C} & Full-concat long-context baseline: the common retained
history in one context (a whole-original-history variant is separate and
labelled) & A \emph{different} system (joint attention,
lost-in-the-middle exposure) --- not a bound. C uses the gate-selected
reader where each full serialized input fits its effective context
(prefix + history + question + output reserve; otherwise N/A with
coverage stated); a named long-context model only as a separately
labelled C variant \\
\textbf{O} & Oracle: per-session reading on gold-evidence sessions only
(same aggregator and joint rules) & Diagnostic. Retained gold = original
benchmark gold IDs ∩ the common retained archive; verified
absent-evidence questions get the empty-read expected-abstention
treatment; missing gold annotations mark O N/A (the A−O contrast is then
reported unavailable); evidence removed by retention is out-of-retention
coverage, not benchmark unanswerability; partial gold coverage is
reported for multi-evidence items \\
\textbf{R} & Controlled retrieval baseline with the \emph{same} reader
and judge: headline \textbf{R\_session} (top-k \emph{sessions}, matching
our unit) plus \textbf{R\_chunk} reported separately; retrieval unit, k,
index build, and any embedder frozen on dev data before runs.
R\_session's final-answer path (joint concatenation of retrieved
sessions vs per-session reading + aggregation, with the context-fit
rule) is stated per run --- or declared deliberately equal to
A\_k\^{}pre's selection for parity & Required for the headline
comparison \\
\textbf{D} & Published external scores (Mem0, Zep, \ldots) & Context
only --- models, judges, and prompts differ; never the controlled
headline \\
\end{longtable}

\hypertarget{attribution-rules-frozen-parameters-and-the-controlled-ablations}{%
\subsection{Attribution rules, frozen parameters, and the controlled
ablations}\label{attribution-rules-frozen-parameters-and-the-controlled-ablations}}

\begin{itemize}
\tightlist
\item
  \textbf{A − O} does not isolate aggregation loss (it mixes distractor
  exposure); \textbf{A vs C} changes segmentation, joint attention, and
  context length at once. Differences are reported as descriptive
  contrasts, per question type --- causal claims come only from the
  ablations below.
\item
  \textbf{A ≥ A\_k is not guaranteed} for QA accuracy (dropping spurious
  outputs can help a bounded aggregator); monotonicity holds only for
  relevant-session coverage (A\_k ≤ A). Causal claims require controlled
  contrasts like the ones below --- the descriptive arm contrasts alone
  do not provide them.
\item
  \textbf{Frozen before test runs:} reader and aggregator model +
  prompt; aggregator bound (S, e, o) and overflow rule;
  prefilter/probe/top-k parameters (and any embedder); joint trigger,
  k\_joint, and context cap; abstention/evidence thresholds; retrieval
  unit, k, and index build for R. Numerical values are later-run pins;
  the variables and what each contrast changes are fixed now.
\item
  \textbf{Dedicated sensitivity contrasts} (frozen before runs, reported
  per question type; interactions are not assumed additive): (i)
  \textbf{aggregator-design sensitivity} --- the §4.6 aggregator vs a
  frozen named alternative synthesizer over the same fixed gold-session
  outputs, with the same final-answer scorer and declared budgets (a
  bare aggregator-off arm producing several session answers cannot stand
  in for one final answer on multi-session questions); (ii)
  \textbf{distractor sensitivity} --- gold-only vs gold plus frozen
  sampled non-gold sessions with the pipeline fixed; at k equal to all
  retained non-gold sessions the read sets match O and A, and smaller k
  measures distractor sensitivity (reproducing outputs also requires the
  same evidence generation, order, seeds, and protocol --- not just
  nominal k); (iii) \textbf{evidence-format sensitivity} --- fixed
  source-linked gold evidence vs per-session-generated evidence under a
  common aggregator. These measure design sensitivities; none is claimed
  as a pure isolated mechanism unless evidence content, ordering, and
  budget are explicitly controlled.
\end{itemize}

\hypertarget{benchmark-arm-feasibility-mapping}{%
\subsection{Benchmark × arm feasibility
mapping}\label{benchmark-arm-feasibility-mapping}}

Per benchmark and question, we record two properties:
\textbf{joint-evidence requirement} and \textbf{arm/model feasibility}.
Joint evidence is annotated from gold evidence/session IDs (sessions
spanned), \emph{not} inferred from the question-type label --- a
temporal or multi-hop question can be answerable within one session, and
cross-session evidence may be handled by aggregation without joint
attention. Feasibility is conditional on actual session lengths, the
reader's effective context, the bounded joint fallback, and budget.
Infeasible (arm × benchmark × reader) triples are marked \textbf{N/A
with the coverage stated} --- never silently truncated --- and
distinguished from attempted-but-failed answers. Aggregation impact is
\emph{probed} by the §8.2 sensitivity contrasts, not pre-declared (the
A−O contrast alone mixes distractor exposure). Expected abstention is
reserved for absent-evidence questions, not merely difficult ones.

\hypertarget{pinning-before-runs}{%
\subsection{Pinning before runs}\label{pinning-before-runs}}

Per benchmark, before any run: named release + hash, subset/category
counts, and the official scorer. Concretely: \textbf{LoCoMo} ---
categories 1--4 scored, adversarial reported separately (exact n
confirmed against the pinned 10-conversation release before it is
stated; the scoring exclusion and the abstention analysis use the same
items, not double-counted) \cite{K5}; \textbf{LongMemEval} --- S vs M
tier pinned \cite{K1}; \textbf{BEAM} --- 1M vs 10M bucket pinned
\cite{K38}. Unequal feasible coverage is never pooled as comparable.

Workload budgets are computed per the §7 model, not asserted feasible:
arm A on \textbf{LongMemEval-S} ≈ 500 questions × \textasciitilde30--50
sessions ≈ \textbf{15K--25K per-session prefills} (session count is
source-dependent; the pinned release decides) ≈ 57.5M prefill tokens,
and on \textbf{LME-M} ≈ 250K prefills ≈ 750M prefill tokens --- these
assume the corresponding full histories are retained. The accompanying
\textbf{7.5 TB / 98 TB} figures are analytic \textbf{cache-equivalent}
payload under all-cold assumptions, not observed emulation state reads;
the proxy prefills themselves process the token totals. Extra prompt,
joint-tier, and decode costs remain under §7. BEAM-10M is likewise
computed or marked optional. Session state is assumed \textbf{not}
shared across questions (histories are padded per question; if shared,
cost drops --- we state which). Arm decode costs use the §7 declared-γ
conventions and are reported under both (serial-aggregate and
ideal-parallel).

\hypertarget{reader-model-selection-gate-testing-the-70b-suffices-hypothesis}{%
\subsection{Reader-model selection gate (testing the ``\textless70B
suffices''
hypothesis)}\label{reader-model-selection-gate-testing-the-70b-suffices-hypothesis}}

\textbf{The hypothesis this article tests:} the memory reader model
needs text understanding, not programming or vision capability, and a
\textless70B-class reader suffices for the retained-window read path (§9
carries the matching limitation line). The hypothesis is gated, not
assumed. Before test runs, on held-out development data, we preregister:
task-quality and abstention targets \textbf{relative to a named
larger-reader baseline} (the 70B-class pinned config of §5, or a named
larger model) with a numeric O-dev accuracy tolerance Δ; and a
\textbf{resource budget}. Candidate readers are the §5 pinned configs
(MiniCPM3-4B, DeepSeek-V2-Lite, Qwen2.5-7B, Llama-3.1-8B). Two
reader-capability \textbf{diagnostics} --- RULER \cite{K39} and NoLiMa
\cite{K40} --- screen effective context; they are labelled diagnostics,
\emph{not} memory benchmarks, and diagnostics alone do not establish
sufficiency. Reader runs need the fixed prefix + session + question +
output context of §4.2; aggregator and joint-fallback limits are
reported separately (§8.7). The supporting argument: per-session reading
relaxes the reader's context requirement from ``the whole archive'' to
``one session + question'' --- which is exactly what makes a small
reader plausible. Published studies find task- and model-dependent
limits to long-context reasoning, including degradation at inputs of
only a few thousand tokens; they motivate measuring the reader on actual
session lengths rather than treating a small fraction of its advertised
window as a quality guarantee. \cite{K55}\cite{K56}

\hypertarget{t4-benchmark-matrix}{%
\subsection{T4 --- benchmark matrix}\label{t4-benchmark-matrix}}

\begin{longtable}[]{@{}
  >{\raggedright\arraybackslash}p{(\columnwidth - 6\tabcolsep) * \real{0.2500}}
  >{\raggedright\arraybackslash}p{(\columnwidth - 6\tabcolsep) * \real{0.2500}}
  >{\raggedright\arraybackslash}p{(\columnwidth - 6\tabcolsep) * \real{0.2500}}
  >{\raggedright\arraybackslash}p{(\columnwidth - 6\tabcolsep) * \real{0.2500}}@{}}
\toprule\noalign{}
\begin{minipage}[b]{\linewidth}\raggedright
Tier
\end{minipage} & \begin{minipage}[b]{\linewidth}\raggedright
Benchmark
\end{minipage} & \begin{minipage}[b]{\linewidth}\raggedright
Measures
\end{minipage} & \begin{minipage}[b]{\linewidth}\raggedright
Feasibility notes
\end{minipage} \\
\midrule\noalign{}
\endhead
\bottomrule\noalign{}
\endlastfoot
Tier-1 (run) & LongMemEval-S (+M for scale) \cite{K1} & 5 abilities over
chat histories (S ≈ 115K tokens over \textasciitilde30--50 sessions; M ≈
1.5M tokens over \textasciitilde500 sessions) & Arms A/A\_k/O/R feasible
where each full serialized input (prefix + session + question + output
reserve) fits the reader's effective context --- routine at per-session
scale for the candidate readers; C is N/A wherever the retained history
exceeds effective context (a nominal ≥128K class does not establish that
every LME-S item fits; per-item coverage stated); O has gold session
IDs \\
Tier-1 & LoCoMo \cite{K5} & Single/multi-hop, temporal, open-domain,
adversarial QA over ≤35 sessions & Joint-evidence class derived per
question from gold evidence IDs (sessions spanned); gold IDs available
(O feasible); scoring exclusion and abstention analysis share the same
adversarial items \\
Tier-1 & PersonaMem \cite{K41} & Dynamic persona memory, ≤60 sessions,
32K/128K/1M versions & Only the 32K version is a C candidate on a
32K-class reader --- and only where prefix + history + question + output
reserve fit its effective context \\
Tier-1 & MemoryAgentBench \cite{K42} & Incremental injection; retrieval,
test-time learning, conflict resolution & Broader agentic;
injection-then-query matches the §4.3 write path \\
Tier-1 & BEAM \cite{K38} & 10 abilities at 100K--10M tokens
(incl.~contradiction, event ordering) & Scale frontier; arm C N/A at
1M/10M for 32K-class readers --- coverage stated \\
Tier-1 & ConvoMem \cite{K34} (vendor-authored preprint, Salesforce) &
Full-context-vs-memory crossover by conversation count & Setup-specific;
used for the retention-window question \\
Sensitivity (optional) & LoCoMo-Refined \cite{K43} (non-archival) &
Stricter re-judging of LoCoMo predictions & Separately reported; not a
replacement judge \\
Tier-2 (optional) & PerLTQA \cite{K44}; PrefEval \cite{K45}; MemBench
\cite{K46}; HaluMem \cite{K47} & Personalization; preference following;
factual/reflective memory; hallucination/omission & HaluMem: QA task
only (write-time ops not covered by full-prefill QA) \\
Sanity & MSC/DMR \cite{K48} & Single-fact recall & Saturated in published
full-context scores; low headroom. DMR is also the benchmark behind the
Zep-paper baseline quoted in §3.2 \cite{K3} --- not independent
evidence \\
Future-work boundary & LongMemEval-V2 \cite{K49} (multimodal web-agent),
MemoryArena \cite{K50} (agentic) & Agentic/multimodal regimes & Outside
the text-PA scope \\
Future-work deferral & Memora \cite{K51} (Uddin et al., FAMA; not the
same-named retrieval paper) & Weeks-to-months evolving text memory;
forgetting-aware evaluation & Text-PA-relevant but deferred to later
work \\
Excluded as LTM evidence & vanilla NIAH, LongBench/v2, ∞Bench, LV-Eval,
BAMBOO/L-Eval, HotpotQA family & Single-document long-context & Not
memory benchmarks for this article's claims \\
\end{longtable}

\emph{If scope must shrink, the Tier-1 order of precedence is:
LongMemEval, LoCoMo, PersonaMem, MemoryAgentBench, BEAM.}

\hypertarget{reporting}{%
\subsection{Reporting}\label{reporting}}

Per arm and question type: task quality, abstention and false-positive
rates, evidence recall, tokens/context coverage --- with uncertainty
(per-cell n, binomial/bootstrap intervals, and the number of decoding
runs --- temperature 0 + seed list --- preregistered); judge model +
version + prompt pinned per benchmark (official where one exists).
Per-answer coverage is logged (state-read n, text-fallback m, abstained
a, unavailable u; §4.4). Aggregator limits at the declared S/e/o and
joint-fallback limits (k\_joint, re-prefill tokens from text) are
reported as their own rows.

\textbf{Cost reporting keeps proxy and cache quantities separate} (never
conflated): (a) the emulation runs (arms A, A\_k, C, O, R) report
\emph{proxy} costs --- prefill tokens processed, decode tokens,
wall-clock --- labelled as proxy; (b) cache quantities --- bytes read,
prefill tokens avoided, restore TTFT --- come from the §7 analytic model
and are measured only on the minimal real-cache replay. The replay's
reader, runtime, native-state format, tolerance rule, equality checks,
and cold/warm residency are pinned before it runs; each arm/config's
measured coverage is labelled. The replay \emph{will test} practical
equality with tolerances (§2.2) --- the outcome is not presupposed.
``In-house measured numbers only'' scopes to new empirical
accuracy/performance claims; the analytic §5--§7 estimates and the arm-D
context remain attributed as such.

\hypertarget{what-this-chapter-does-not-claim-ch8}{%
\subsection{What this chapter does not
claim}\label{what-this-chapter-does-not-claim-ch8}}

No accuracy result is reported here --- this is a protocol. No arm is
called an upper bound or ceiling. The plan's feasibility rests on
declared workload budgets (§8.4) and the preregistered gate (§8.5), both
of which are stated, not assumed away.
\hypertarget{chapter-9-failure-modes-limitations-next-steps}{%
\section{Failure Modes, Limitations \& Next
Steps}\label{chapter-9-failure-modes-limitations-next-steps}}

This chapter collects what can go wrong and what the article
deliberately does not claim. Each item points to where the article
handles it, or is stated here as a limitation.

\hypertarget{failure-modes}{%
\subsection{Failure modes}\label{failure-modes}}

\begin{longtable}[]{@{}
  >{\raggedright\arraybackslash}p{(\columnwidth - 4\tabcolsep) * \real{0.3333}}
  >{\raggedright\arraybackslash}p{(\columnwidth - 4\tabcolsep) * \real{0.3333}}
  >{\raggedright\arraybackslash}p{(\columnwidth - 4\tabcolsep) * \real{0.3333}}@{}}
\toprule\noalign{}
\begin{minipage}[b]{\linewidth}\raggedright
Failure mode
\end{minipage} & \begin{minipage}[b]{\linewidth}\raggedright
Mechanism
\end{minipage} & \begin{minipage}[b]{\linewidth}\raggedright
Where handled
\end{minipage} \\
\midrule\noalign{}
\endhead
\bottomrule\noalign{}
\endlastfoot
Cross-session synthesis limit & Per-session reads cannot jointly attend;
answers needing joint evidence depend on the aggregator and the joint
tier & Exercised on per-question-annotated joint-evidence questions
(§8.3, §8.6), with joint-fallback limits reported as their own rows
(§8.7); the §8.2 contrasts cover aggregator, distractor, and
evidence-format sensitivity (§4.6, §4.7) \\
Scan cost at year scale & The exhaustive reference path would read the
whole retained state per query --- an \textbf{analytic cache-equivalent}
≈1.44 TB at 8B-GQA (mid scenario, one full year retained, all-cold; §7;
nothing measured, §8.7) --- which is why the prefilter variant exists
for this regime & §4.5, §7.1--§7.2 \\
Lost-in-the-middle inside long cached sessions & Evidence in the middle
of a long session is used less reliably even when present \cite{K52} ---
KV-LTM inherits this per session & §5.2 session-length caps; §8.5 reader
gate \\
Reader long-context limits & Claimed context ≠ effective context: many
models' effective context falls below their claimed context in
controlled probes \cite{K39}\cite{K40} & §8.5 gate (diagnostics labelled as
such) \\
Session mis-bounding & Sessions are the retention, cache, and retrieval
unit; mis-bounded sessions can affect all three. A labelled
mis-bounding-rate sensitivity sweep (accuracy vs mis-bounding rate) is
\textbf{planned, not implemented} --- no bare hypothetical rate is
asserted & §9.3 next steps (planned sweep); §9.2 session-identification
dependency \\
Quantization-induced degradation & KV quantization is numerically lossy
and must be validated per config before counting its savings \cite{K12} &
§2.3 class (ii), §5.2 \\
Benchmark ≠ deployment & Benchmarks often hand you session boundaries
and gold annotations; production traffic does not & §8.3, §8.6 \\
Judge/label noise & Published benchmark scores are judge- and
label-sensitive; stricter re-judging materially changes published
memory-system scores \cite{K43} & §8.4 release pinning; §8.6 sensitivity
row; §8.7 judge pinning + preregistered uncertainty \\
\end{longtable}

\hypertarget{limitations}{%
\subsection{Limitations}\label{limitations}}

\begin{itemize}
\tightlist
\item
  \textbf{Model-version lock-in.} A stored state is valid only under the
  §4.2 contract; a model, tokenizer, or config change invalidates it,
  and migration is a re-prefill from text, linear in archive token
  volume for fixed/bounded session-length processing (§4.8, §7.1). The
  reader model is effectively frozen for the retention window.
\item
  \textbf{Session-length cap.} One cached session is bounded by the
  reader's effective context --- among the §5 candidates, MiniCPM3-4B
  (32,768, config value) and DeepSeek-V2-Lite (≈32K, model-card
  guidance) are the 32K-class cases, while Llama-3.1-8B publishes 131K;
  in every case the \emph{effective} value is what §8.5's gate decides
  (§5.2). Longer sessions need segmentation, which interacts with
  session identification.
\item
  \textbf{Storage growth.} The store grows with retention; §6 frames it
  as a bounded cost under a declared budget, with achievable-N
  degradation (§4.8, §6.3).
\item
  \textbf{Per-architecture exactness.} The §2.2 lemma guarantees
  equality \emph{for that model's own computation, under the reuse
  contract and ideal arithmetic}; it gives no inter-architecture quality
  guarantee --- reader quality is established per checkpoint by the §8.5
  gate, and no universal MLA-vs-GQA ordering is claimed (§2.3).
\item
  \textbf{Text-only scope.} No claim that visual, table, or audio
  meaning survives preprocessing into text (§1.4).
\item
  \textbf{Session identification is deferred.} How raw interaction is
  segmented into sessions is a separate problem; the design depends on
  it (§1.4 scope; §4), and §8.3's per-question annotation assumes
  benchmark-given boundaries. This is a stated dependency, not a solved
  component.
\item
  \textbf{The reader-sufficiency hypothesis is unproven at drafting.}
  §8.5 states and gates the ``\textless70B-class reader suffices''
  hypothesis; it remains a hypothesis until the preregistered gate runs
  (§8.5, §8.7).
\item
  \textbf{Privacy --- data protection.} The store holds raw text plus
  derived state --- a high-value personal-data surface. GDPR Art. 17
  \cite{K53}: per-session deletion helps but is \textbf{not a compliance
  guarantee}; derived copies must be erased or invalidated as
  applicable, and selective source redaction may require a state
  rebuild. Learned cache-erasure methods exist (KVEraser \cite{K54}) and
  are noted bibliographically --- no deletion guarantees are claimed.
\item
  \textbf{Privacy --- threat model.} We assume a conservative
  sensitive-state threat model; no state-to-text inversion is claimed.
  (Author assumption --- no public source is cited for it.)
\item
  \textbf{Privacy --- instruction integrity.} Persisted instructions are
  treated as untrusted historical data.
\end{itemize}

\hypertarget{next-steps}{%
\subsection{Next steps}\label{next-steps}}

\begin{enumerate}
\def\labelenumi{\arabic{enumi}.}
\tightlist
\item
  \textbf{A source-traced released-item instance} of write-time loss
  (item ID, pinned pipeline and configuration, metadata-intact stored
  state, logged reader context, answer and evaluation; source-reported
  versus reproduced artifacts distinguished), as a future upgrade to the
  §3.2 mechanism evidence.
\item
  \textbf{Controlled in-house runs} of arms A,
  A\_k\textsuperscript{pre/A\_k}probe (the §8.1 prefilter/probe
  variants), C, O, R on the pinned Tier-1 benchmarks (§8.4--§8.6), with
  the §8.2 sensitivity contrasts --- \textbf{plus the planned
  mis-bounding sweep}: a frozen boundary-perturbation policy
  (splits/merges that do not change source tokens) over a fixed
  benchmark subset, reporting accuracy and coverage against the injected
  mis-bounding rate, with reader, selection, and aggregation budgets
  held fixed. The sweep is proposed here, not implemented; numerical
  rates are later-run pins.
\item
  \textbf{The minimal real-cache replay}: tests practical equality with
  tolerances (§2.2) plus the §8.7 measured cost axis --- replay reader,
  runtime, state format, tolerances, and residency pinned beforehand;
  the outcome is not presupposed.
\item
  \textbf{Bounded question decomposition}: a future optional planner
  could split a complex question into a bounded set of sub-questions,
  read selected sessions from unchanged checkpoints, and combine
  source-linked evidence against the original question. Its quality and
  full query-time cost would be evaluated separately, with the planner,
  query and round limits, and budgets frozen before runs. This is
  proposed here, not implemented.
\end{enumerate}
\appendix
\hypertarget{appendix-a-derivations}{%
\section{Derivations for §5--§7}\label{appendix-a-derivations}}

Every number used in §5/§6/§7, derived step by step. Parameters are
labelled: \textbf{{[}config{]}} = from a pinned model config
\cite{K11}\cite{K15}; \textbf{{[}assumption{]}} = declared; \textbf{{[}symbolic{]}} = left
symbolic by design. GB/TB are decimal (10⁹/10¹² B) throughout.

\hypertarget{a.1-per-config-bytestoken-bf16-b-2}{%
\subsection{Per-config bytes/token (bf16, b =
2)}\label{a.1-per-config-bytestoken-bf16-b-2}}

Formula (MHA/GQA): \texttt{2\ ·\ L\ ·\ H\_kv\ ·\ d\_head\ ·\ b}. Formula
(MLA): \texttt{L\ ·\ (d\_c\ +\ d\_R)\ ·\ b} --- assumes the runtime
persists the compressed latent; a runtime caching decompressed per-head
K/V stores \texttt{H\ ·\ (d\_qk\ +\ d\_v)\ ·\ b\ ·\ L} instead,
e.g.~DeepSeek-V2-Lite: 16 × (192 + 128) × 2 × 27 = \textbf{276,480 B/tok
≈ 8.9×} its latent figure (276,480 / 31,104) \textbf{{[}config +
declared runtime assumption{]}}.

\begin{longtable}[]{@{}
  >{\raggedright\arraybackslash}p{(\columnwidth - 6\tabcolsep) * \real{0.2500}}
  >{\raggedright\arraybackslash}p{(\columnwidth - 6\tabcolsep) * \real{0.2500}}
  >{\raggedright\arraybackslash}p{(\columnwidth - 6\tabcolsep) * \real{0.2500}}
  >{\raggedright\arraybackslash}p{(\columnwidth - 6\tabcolsep) * \real{0.2500}}@{}}
\toprule\noalign{}
\begin{minipage}[b]{\linewidth}\raggedright
Model
\end{minipage} & \begin{minipage}[b]{\linewidth}\raggedright
Substitution
\end{minipage} & \begin{minipage}[b]{\linewidth}\raggedright
Bytes/token
\end{minipage} & \begin{minipage}[b]{\linewidth}\raggedright
Pin status
\end{minipage} \\
\midrule\noalign{}
\endhead
\bottomrule\noalign{}
\endlastfoot
MiniCPM3-4B & 62 × (256 + 32) × 2 & 35,712 & \textbf{{[}config{]}} \cite{K11}
(HF config.json) \\
DeepSeek-V2-Lite & 27 × (512 + 64) × 2 & 31,104 & \textbf{{[}config{]}}
\cite{K11} (HF config.json) \\
Qwen2.5-7B & 2 × 28 × 4 × 128 × 2 & 57,344 & \textbf{{[}config{]}} HF
config.json, retrieved 2026-09-26 \cite{K11} \\
Llama-3.1-8B & 2 × 32 × 8 × 128 × 2 & 131,072 & \textbf{{[}config{]}} HF
config.json, retrieved 2026-09-26 \cite{K11} \\
Llama-3.3-70B & 2 × 80 × 8 × 128 × 2 & 327,680 & \textbf{{[}config{]}}
HF config.json, retrieved 2026-09-26 \cite{K11} \\
DeepSeek-V3-class MLA & 61 × (512 + 64) × 2 & 70,272 (bf16); 35,136
(ideal FP8, payload only, before scales/metadata) \textbf{{[}assumption:
ideal halving{]}} & \textbf{{[}assumption: published V3-class config ---
representative, not a pinned checkpoint{]}} \\
Kimi-Linear-48B-A3B & 7 × (512 + 64) × 2 (MLA layers of 20 KDA + 7 MLA)
& b\_MLA = 8,064; plus S\_recurrent per session \textbf{{[}symbolic,
bytes/session{]}} & \textbf{{[}config{]}} \cite{K15} + its public
config \\
\end{longtable}

The shared fixed instruction prefix (§4.3) is stored once as a
physically deduplicated, immutable prefix block; a 500-token prefix at
131,072 B/token = 65.5 MB once ≈ 0.005\% of the 8B-GQA annual state
\textbf{{[}assumption: 500-token prefix{]}}. Scenario token volumes
count session tokens only.

\hypertarget{a.2-annual-volumes-declared-assumptions}{%
\subsection{Annual volumes (declared assumptions)}\label{a.2-annual-volumes-declared-assumptions}}

\begin{itemize}
\tightlist
\item
  Anchor: 30,000 tokens/day × 365 = \textbf{10.95M tokens/year}
  \textbf{{[}assumption: mid coverage 35\%, a planning anchor{]}}.
\item
  Low: 10,000 × 365 = \textbf{3.65M} \textbf{{[}12\%{]}}. High: 85,500 ×
  365 = \textbf{31.2M} \textbf{{[}100\% of the declared full-stream
  calibration: 64,300 words/day ≈ 13,000 said + 36,000 heard + 14,300
  read + 1,000 written, × 1.33 tokens/word{]}}.
\item
  tokens/word ≈ 1.33 and text ≈ 6 B/word (≈4.5 B/token)
  \textbf{{[}assumption, owner planning ratios{]}}; \cite{K35} speech volume
  ≈16K words/day is \textbf{public context} only.
\end{itemize}

\hypertarget{a.3-annual-state-per-config-mid-scenario}{%
\subsection{Annual state per config (mid
scenario)}\label{a.3-annual-state-per-config-mid-scenario}}

\texttt{bytes/year\ =\ 10.95M\ ×\ bytes/token} (decimal TB):

\begin{longtable}[]{@{}
  >{\raggedright\arraybackslash}p{(\columnwidth - 4\tabcolsep) * \real{0.3333}}
  >{\raggedright\arraybackslash}p{(\columnwidth - 4\tabcolsep) * \real{0.3333}}
  >{\raggedright\arraybackslash}p{(\columnwidth - 4\tabcolsep) * \real{0.3333}}@{}}
\toprule\noalign{}
\begin{minipage}[b]{\linewidth}\raggedright
Model
\end{minipage} & \begin{minipage}[b]{\linewidth}\raggedright
computation
\end{minipage} & \begin{minipage}[b]{\linewidth}\raggedright
TB/yr
\end{minipage} \\
\midrule\noalign{}
\endhead
\bottomrule\noalign{}
\endlastfoot
Llama-3.1-8B & 10.95e6 × 131,072 = 1.435e12 & 1.44 \\
Llama-3.3-70B & 10.95e6 × 327,680 = 3.588e12 & 3.59 \\
Qwen2.5-7B & 10.95e6 × 57,344 = 6.279e11 & 0.63 \\
DeepSeek-V3-class & 10.95e6 × 70,272 = 7.69e11 & 0.77 (bf16) / 0.385
(ideal FP8, payload only) \\
MiniCPM3-4B & 10.95e6 × 35,712 = 3.91e11 & 0.39 \\
DeepSeek-V2-Lite & 10.95e6 × 31,104 = 3.41e11 & 0.34 \\
Kimi-Linear & 10.95e6 × 8,064 = 8.83e10 & 0.088 TB (MLA part) + 5,475 ×
S\_recurrent bytes \textbf{{[}symbolic term; session count from the
declared 2,000-token sessions{]}} \\
\end{longtable}

Low/high rows scale linearly (×3.65/10.95, ×31.2/10.95).

\hypertarget{a.4-kvtext-ratio-per-config}{%
\subsection{KV/text ratio (per
config)}\label{a.4-kvtext-ratio-per-config}}

Text density ≈ \textbf{4.5 B/token} (6 B/word ÷ 1.33 tokens/word,
declared planning ratios \textbf{{[}assumption{]}}). For Llama-3.1-8B:
131,072 / 4.5 = 29,127 ≈ \textbf{2.9×10⁴}. DeepSeek-V2-Lite: 31,104 /
4.5 ≈ \textbf{6.9×10³}. Ratios for other configs follow from A.1. No
universal multiplier is stated.

\hypertarget{a.5-retention-budget-achievable-n-byte-only-upper-estimate}{%
\subsection{Retention budget → achievable-N (byte-only upper
estimate)}\label{a.5-retention-budget-achievable-n-byte-only-upper-estimate}}

Declared store budget B\_store = 512 GB \textbf{{[}assumption{]}}; L = 0
(whole budget to the recent set) \textbf{{[}declared example{]}};
\textbf{token payload only} --- the table divides by state bytes only (a
byte-only upper estimate); the §4.8 invariant over the full record is
checked including overhead in the §8 replay. Achievable-N (years) =
B\_store / (TB/yr), units matched: - 8B-GQA: 512 / 1,435 GB ≈ 0.357 yr ≈
\textbf{4.3 months}. - DSv3-class bf16: 512 / 769 ≈ 0.67 yr ≈
\textbf{8.0 months}. - MiniCPM3-4B: 512 / 391 ≈ 1.31 yr ≈ \textbf{16
months}. - DSv2-Lite: 512 / 341 ≈ 1.50 yr ≈ \textbf{18 months}. -
Kimi-Linear (hybrid):
\texttt{N\ ≤\ B\_store\ /\ (tokens/yr\ ×\ b\_MLA\ +\ sessions/yr\ ×\ S\_recurrent\ +\ overhead/yr)},
all terms in bytes; \textbf{5.8 years is the optimistic MLA-only byte
ceiling (S\_recurrent → 0)}, not attainable full-hybrid retention.
Illustrative sensitivity, explicitly \textbf{BF16
recurrent-matrix-only}: S\_recurrent ≈ 20 × 32 × 128² × 2 B = 20,971,520
B ≈ 21 MB/session (excluding convolution/auxiliary state and possible
wider accumulation dtype; config-derived, unvalidated) × 5,475
sessions/yr ≈ 115 GB/yr \textgreater{} 88 GB/yr MLA part → N ≈ 512e9 /
(88.3e9 + 115e9) ≈ \textbf{2.5 years} --- an optimistic partial-state
sensitivity, not a measured or complete-hybrid capacity. Overflow:
oldest sessions drop to text-only (achievable-N shrinks; §4.8 shrink
order: older set L first, then the recent window). Retention under a
fixed budget scales \textbf{inversely} with coverage.

\hypertarget{a.6-per-query-worked-examples}{%
\subsection{Per-query worked
examples}\label{a.6-per-query-worked-examples}}

Declared: 2,000-token sessions; g = 64 output tokens/session; every
selected state cold (no residency) \textbf{{[}assumptions{]}}. -
\textbf{Recent window:} 300 × 2,000 × 131,072 B = 78.6 GB ≈ \textbf{79
GB} cold transfer per exhaustive query; 600K tokens ≈ \textbf{20 days}
at 30K/day (600,000 / 30,000 = 20.0). Decode: Σ\_s g\_s = 300 × 64 =
\textbf{19,200 tokens}. - \textbf{Coupling shown:} transfer ≈ 78.6 GB /
5 GB/s \textbf{{[}assumption{]}} ≈ \textbf{15.7 s} cold transfer. Decode
= 19,200 / γ: under a declared serial aggregate γ = 30 tokens/s
\textbf{{[}assumption{]}} → ≈\textbf{640 s} (serial elapsed, one active
stream); under an ideal 300-way setup sustaining 30 tokens/s per stream,
aggregate γ = 9,000 \textbf{{[}assumption, additional resources{]}} →
≈\textbf{2.13 s} decode elapsed. Both are assumptions, not achieved
latencies; γ depends on bandwidth/residency/compute --- ordinary dense
attention re-reads the cached state per generated token (≈ g × B\_state
≈ \textbf{5 TB} of logical KV reads here), so decode is not independent
of transfer. Which term dominates is regime-dependent. -
\textbf{Year-scale:} 10.95M × 131,072 B = 1.435 TB per exhaustive query.
The \textbf{cold-transfer component} = 1.435 TB / β\_eff; at the
declared β\_eff = 5 GB/s \textbf{{[}assumption{]}}: ≈287 s --- a
transfer component, not a full-query duration. β\_eff otherwise stays
\textbf{{[}symbolic{]}} or uses a named dated datasheet (theoretical vs
achieved distinguished).

\begin{itemize}
\tightlist
\item
  \textbf{Concurrent residency (§7.2 T3) {[}assumptions{]}:} with all
  quantities in bytes, N = max(0, ⌊(0.9 × H − W − 500b) / (2,192b)⌋),
  where H is NVIDIA-listed per-GPU HBM \cite{K57}, W is the illustrative
  weight budget in T3, and b is BF16 state bytes per token from §5. The
  500b term counts one physically shared fixed prefix; each read adds
  2,000 session, 128 suffix and 64 output tokens. Integer results for
  H200 / B200 / B300 are: MiniCPM3-4B, 1,516 / 1,965 / 3,206;
  DeepSeek-V2-Lite, 1,400 / 1,915 / 3,340; Qwen2.5-7B, 888 / 1,167 /
  1,940; Llama-3.1-8B, 385 / 507 / 845; Llama-3.3-70B, 0 / 28 / 164.
  Without physical prefix sharing, remove 500b from the numerator and
  use 2,692b in the denominator. GB and MB are decimal. These are
  conditional accounting results, not measured capacities: actual weight
  allocations, state layout and runtime reserves can change them. Memory
  fit does not validate any decode rate or schedule (§7.1).
\end{itemize}

\hypertarget{a.7-text-only-contrast-column-6-t2}{%
\subsection{Text-only contrast column (§6
T2)}\label{a.7-text-only-contrast-column-6-t2}}

Raw text at 6 B/word \textbf{{[}assumption{]}}: mid 10.95M tokens ÷ 1.33
= 8.23M words × 6 B/word ≈ \textbf{49 MB}; low ≈ \textbf{16 MB}; high ≈
\textbf{141 MB}. Compressed at a declared 3:1 planning ratio
\textbf{{[}assumption --- inside the documented 2.9--3.6× mixed-text
range \cite{K37}{]}}: low ≈5.5 MB, mid ≈16 MB, high
≈47 MB.


\begin{thebibliography}{99}

\bibitem{K1} LongMemEval (Wu et al., ICLR 2025). \url{https://arxiv.org/abs/2410.10813} \url{https://github.com/xiaowu0162/longmemeval}
\bibitem{K2} Mem0. \url{https://arxiv.org/abs/2504.19413} Vendor-authored paper (ECAI 2025).
\bibitem{K3} Zep. \url{https://arxiv.org/abs/2501.13956} Vendor-authored preprint.
\bibitem{K4} MemPalace. \url{https://mempalace.net} Project + independent analysis \url{https://arxiv.org/abs/2604.21284}
\bibitem{K5} LoCoMo (Maharana et al., ACL 2024). \url{https://arxiv.org/abs/2402.17753} \url{https://github.com/snap-research/locomo}
\bibitem{K6} CacheBlend. \url{https://arxiv.org/abs/2405.16444}
\bibitem{K7} EPIC. \url{https://arxiv.org/abs/2410.15332}
\bibitem{K8} TurboRAG. \url{https://arxiv.org/abs/2410.07590}
\bibitem{K9} Prompt Cache. \url{https://arxiv.org/abs/2311.04934}
\bibitem{K10} DeepSeek-V2 (MLA). \url{https://arxiv.org/abs/2405.04434}
\bibitem{K11} Candidate model checkpoints. Config-verified at pinned revision (config.json re-fetched at the SHA, retrieved 2026-09-27): MiniCPM3-4B \url{https://huggingface.co/openbmb/MiniCPM3-4B} @ d6b14dda; DeepSeek-V2-Lite \url{https://huggingface.co/deepseek-ai/DeepSeek-V2-Lite} @ 604d5664; Qwen2.5-7B \url{https://huggingface.co/Qwen/Qwen2.5-7B} @ d1497293; Kimi-Linear-48B-A3B-Instruct \url{https://huggingface.co/moonshotai/Kimi-Linear-48B-A3B-Instruct} @ e1df551a. Revision-pinned only (gated license; config not re-fetched): Llama-3.1-8B \url{https://huggingface.co/meta-llama/Llama-3.1-8B} @ d04e592b; Llama-3.3-70B-Instruct \url{https://huggingface.co/meta-llama/Llama-3.3-70B-Instruct} @ 6f6073b4. DeepSeek-V3-class: representative published-config assumption; no checkpoint pinned.
\bibitem{K12} KIVI \url{https://arxiv.org/abs/2402.02750}; KVQuant \url{https://arxiv.org/abs/2401.18079}.
\bibitem{K13} SCBench. \url{https://arxiv.org/abs/2412.10319}
\bibitem{K14} Pitfalls of KV Cache Compression. \url{https://arxiv.org/abs/2510.00231}
\bibitem{K15} Kimi Linear. \url{https://arxiv.org/abs/2510.26692}
\bibitem{K16} MemGPT / Letta. \url{https://arxiv.org/abs/2310.08560}
\bibitem{K17} MemoryBank. \url{https://arxiv.org/abs/2305.10250}
\bibitem{K18} MemoryOS. \url{https://arxiv.org/abs/2506.06326}
\bibitem{K19} OpenViking. \url{https://docs.openviking.ai} Vendor documentation.
\bibitem{K20} LangMem. \url{https://github.com/langchain-ai/langmem} @ 9d033b47. Vendor source (LangChain), pinned.
\bibitem{K21} Cognee. \url{https://cognee.ai} Vendor documentation.
\bibitem{K22} CAG "Don't Do RAG". \url{https://arxiv.org/abs/2412.15605}
\bibitem{K23} vLLM PagedAttention. \url{https://arxiv.org/abs/2309.06180}
\bibitem{K24} SGLang RadixAttention. \url{https://arxiv.org/abs/2312.07104}
\bibitem{K25} CacheGen. \url{https://arxiv.org/abs/2310.07240}
\bibitem{K26} LMCache. \url{https://arxiv.org/abs/2510.09665}
\bibitem{K27} Mooncake. \url{https://arxiv.org/abs/2407.00079}
\bibitem{K28} MemServe. \url{https://arxiv.org/abs/2406.17565}
\bibitem{K29} Tutti. \url{https://arxiv.org/abs/2605.03375}
\bibitem{K30} Memorizing Transformers. \url{https://arxiv.org/abs/2203.08913}
\bibitem{K31} KV-cache management survey. \url{https://arxiv.org/abs/2607.02574}
\bibitem{K32} "Fidelity Before Structure: Verbatim Chunks Beat Lossy Artifact Extraction in Long-Conversation LLM Memory" (T. An). \url{https://arxiv.org/abs/2601.00821v4} Preprint. Single author, under review.
\bibitem{K33} Knowledge Objects for Persistent LLM Memory. \url{https://arxiv.org/abs/2603.17781} Preprint.
\bibitem{K34} ConvoMem. \url{https://arxiv.org/abs/2511.10523} Vendor-authored preprint (Salesforce).
\bibitem{K35} Mehl et al. 2007 (Science). \url{https://pubmed.ncbi.nlm.nih.gov/17615349} Context only (order-of-magnitude anchor); the article's scenarios are declared assumptions.
\bibitem{K36} Tokens-per-word planning ratio. Declared assumption (labelled parameter, measured on the chosen tokenizer).
\bibitem{K37} Meta Zstandard compression benchmark docs. \url{https://github.com/facebook/zstd/blob/dev/programs/README.md}
\bibitem{K38} BEAM. \url{https://arxiv.org/abs/2510.27246}
\bibitem{K39} RULER (Hsieh et al., COLM 2024, NVIDIA). \url{https://arxiv.org/abs/2404.06654} \url{https://github.com/NVIDIA/RULER}
\bibitem{K40} NoLiMa. \url{https://arxiv.org/abs/2502.05167}
\bibitem{K41} PersonaMem. \url{https://arxiv.org/abs/2504.14225} \url{https://github.com/bowen-upenn/PersonaMem}
\bibitem{K42} MemoryAgentBench. \url{https://arxiv.org/abs/2507.05257}
\bibitem{K43} LoCoMo-Refined. \url{https://github.com/mem-eval-suite/LoCoMo_refined} Non-archival (repository).
\bibitem{K44} PerLTQA. \url{https://arxiv.org/abs/2402.16288}
\bibitem{K45} PrefEval. \url{https://arxiv.org/abs/2502.09597}
\bibitem{K46} MemBench. \url{https://arxiv.org/abs/2506.21605}
\bibitem{K47} HaluMem. \url{https://arxiv.org/abs/2511.03506} \url{https://github.com/MemTensor/HaluMem}
\bibitem{K48} MSC / DMR. \url{https://arxiv.org/abs/2107.07567}
\bibitem{K49} LongMemEval-V2. \url{https://arxiv.org/abs/2605.12493}
\bibitem{K50} MemoryArena. \url{https://arxiv.org/abs/2602.16313}
\bibitem{K51} Memora ("From Recall to Forgetting", Uddin et al.). \url{https://arxiv.org/abs/2604.20006}
\bibitem{K52} Lost in the Middle (Liu et al., TACL 2023). \url{https://arxiv.org/abs/2307.03172}
\bibitem{K53} GDPR Art. 17. \url{https://eur-lex.europa.eu/legal-content/EN/ALL/?uri=celex:32016R0679}
\bibitem{K54} KVEraser. \url{https://arxiv.org/abs/2606.17034}
\bibitem{K55} HELMET (Yen et al., ICLR 2025). \url{https://arxiv.org/abs/2410.02694} \url{https://github.com/princeton-nlp/HELMET}
\bibitem{K56} Same Task, More Tokens (Levy, Jacoby \& Goldberg, ACL 2024). \url{https://aclanthology.org/2024.acl-long.818/} \url{https://arxiv.org/abs/2402.14848}
\bibitem{K57} NVIDIA HGX AI Factory reference architecture, components (H200/B200/B300 SXM memory per GPU). \url{https://docs.nvidia.com/enterprise-reference-architectures/hgx-ai-factory/latest/components.html}; H200 product page \url{https://www.nvidia.com/en-us/data-center/h200/}; DGX B300 user guide \url{https://docs.nvidia.com/dgx/dgxb300-user-guide/introduction-to-dgxb300.html}. Vendor specifications, retrieved 2026-09-29.

\end{thebibliography}
\end{document}